\documentclass{article}
\usepackage[preprint]{colm2026_conference}
\usepackage[T1]{fontenc}
\usepackage{microtype}
\usepackage{amsmath,amssymb,booktabs,tabularx,longtable,array}
\usepackage{graphicx,tikz}
\usepackage{fontawesome}
\usepackage{xcolor}
\usepackage{placeins}
\usepackage{float}
\usepackage[colorlinks=true,allcolors=blue!45!black,breaklinks=true]{hyperref}
\usepackage{url}
\usetikzlibrary{arrows.meta,calc,positioning,backgrounds}
\definecolor{jink}{HTML}{18212D}
\definecolor{jgray}{HTML}{56606C}
\definecolor{jline}{HTML}{D8DEE7}
\definecolor{jslate}{HTML}{465A78}
\definecolor{jslatelight}{HTML}{E7ECF4}
\definecolor{jgold}{HTML}{8A6508}
\definecolor{jgoldlight}{HTML}{FFF0C2}
\definecolor{jgreen}{HTML}{13755B}
\definecolor{jgreenlight}{HTML}{DDF3E9}
\definecolor{jgraylight}{HTML}{F1F2F4}
\definecolor{jpaper}{HTML}{FFFFFF}
\tikzset{
  every picture/.style={x=1cm,y=-1cm,font=\fontsize{8.6}{10.5}\selectfont,text=jink},
  box/.style={draw=jgray,fill=jgraylight,rounded corners=1.5pt,
    minimum height=.60cm,inner xsep=3.5pt,inner ysep=4pt,align=center,line width=.5pt},
  data/.style={box},
  model/.style={box,draw=jgreen,fill=jgreenlight},
  signal/.style={box,draw=jgold,fill=jgoldlight},
  op/.style={box,draw=jslate,fill=jslatelight},
  stage/.style={minimum height=.95cm},
  tok/.style={box,minimum height=.4cm,minimum width=.5cm,inner sep=2pt},
  flow/.style={-{Stealth[length=1.4mm,width=1mm]},draw=jgray,line width=.65pt,
    rounded corners=1.5pt},
  target/.style={flow,draw=jgold,densely dotted,line width=.9pt},
  grad/.style={flow,dashed,line width=.75pt},
  title/.style={anchor=west,font=\fontsize{9.2}{11.5}\selectfont\bfseries},
  edge/.style={font=\fontsize{7.8}{9.5}\selectfont,text=jgray,inner sep=2pt},
  note/.style={anchor=west,text=jgray,font=\fontsize{8}{9.5}\selectfont},
  smallnote/.style={anchor=west,text=jgray,font=\fontsize{7.6}{9}\selectfont},
  lab/.style={anchor=west},
}

\newcounter{takeaway}
\newcommand{\takeaway}[2]{\refstepcounter{takeaway}\label{#1}\paragraph{Takeaway~\thetakeaway. #2}}
\newcounter{observation}
\newcommand{\observation}[2]{\refstepcounter{observation}\label{#1}\paragraph{Obs.~\theobservation. #2}}
\newcolumntype{L}[1]{>{\raggedright\arraybackslash}p{#1}}
\title{Understanding Jev-Style System One Models:\\[5pt]
  \large Does Structure Reduce Selection Bias?}
\author{Mingyang Song, Mao Zheng, Zhiyuan Ma, Xinpeng Zhou\\
Foundation Model Department, Tencent, China\\
\texttt{nickmysong@tencent.com}\\[2pt]
\faGithub\ \href{https://github.com/nick7nlp/Jev-Selection-Bias}{nick7nlp/Jev-Selection-Bias}}
\date{}
\begin{document}
\maketitle
\lhead{Work in Progress}
\raggedbottom
\begin{abstract}
Selection bias in large language models (LLMs) can persist despite instructions to ignore presentation order or prompts that elicit reasoning. Recent efforts to reproduce Jev-style System One behavior have produced a growing family of models that predict probabilities directly over supplied candidates. Yet systematic comparisons of selection bias across Jev versions, Jev reproductions, and generative LLMs remain limited in multi-option tasks. This leaves a central question: does Jev-style decision structure reduce selection bias? We address this question with a simple test that shuffles options while keeping question and candidate content fixed. Across two Jev versions, twelve Jev reproductions, and six generative LLMs, we trace individual answers through eight presentations and compare them with fixed-order repetition. On our primary collection of 1,700 questions from two benchmarks, Jev's accuracy varies by less than two percentage points, yet its cumulative failure set expands by 27.3--28.5\% relative to the original order, with first failures continuing into the final presentation. Reordering loses substantially more initial successes than fixed-order repetition, a gap that also appears in highly repeatable Jev reproductions. Nor does Jev show a general advantage over the LLM baselines in preserving correct choices. Aggregation offers limited protection: voting yields modest gains for Jev, while even unanimous answers are frequently wrong. These results expose a gap between repeatable decisions and decisions that remain correct across equivalent presentations, showing why stable accuracy alone cannot establish resistance to selection bias.
\end{abstract}

\begin{figure}[H]
\centering
\includegraphics[width=\linewidth]{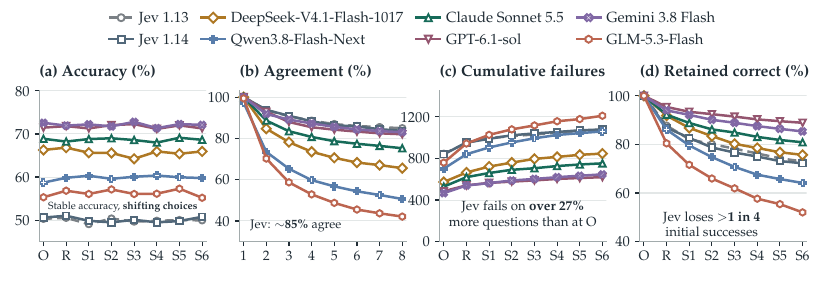}
\caption{\textbf{Stable Jev accuracy hides an expanding failure set.} Two Jev versions and six generative LLMs on 1,700 questions: (a) accuracy, (b) cumulative same-candidate agreement under eight original-order requests, (c) cumulative failed questions, and (d) retention of initial successes. Panels (a,c,d) use the original order (O), reversal (R), and six random permutations (S1--S6). Each failed question counts once; all requests retain the full menu.}
\label{fig:order-overview}
\end{figure}

\clearpage
\lhead{}
\begin{figure}[!htbp]
\centering
\includegraphics[width=\linewidth]{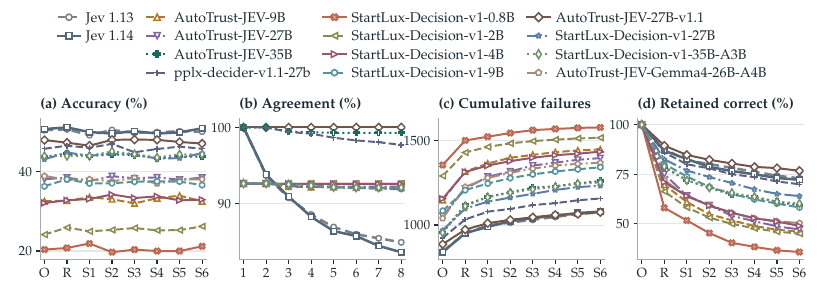}
\caption{\textbf{High repeatability does not ensure order robustness.} Twelve Jev reproductions and both Jev versions on the same 1,700 questions. Panels follow Figure~\ref{fig:order-overview}. The reproductions achieve 91.9--100\% same-candidate agreement under original-order repetition but lose 23.2--64.2\% of initial successes under reordering.}
\label{fig:open-order-overview}
\end{figure}
\section{Introduction}
\label{sec:bias-introduction}

A correct choice should depend on candidate content, yet language models can change their answers when only the presentation changes. Reordering options can alter multiple-choice answers \citep{optionsorder,mcqselectors}, and swapping two unchanged responses can reverse an LLM judge's preference \citep{fairevaluators,mtbench,manyshotevaluators}. Such selection bias lets presentation influence a decision without adding evidence about which candidate is correct. A question answered correctly in one order may therefore fail in another.

Prompting alone does not reliably remove this bias. \citet{mcqselectors} find residual bias under explicit debiasing instructions and chain-of-thought prompting in the settings tested. Calibration and targeted training offer ways to reduce it \citep{calibraeval,symbolbinding,pagrpo}, while evidence from multi-document question answering suggests that reasoning models can be less position-biased \citep{positioncomprehensive}. Selection bias is therefore sensitive to how a model is trained and used. This raises the possibility that models designed expressly for candidate selection could better preserve correct choices across presentations.

Jev-style System One models make this possibility concrete. They predict a probability distribution directly over candidates supplied with each request. TypeSafe describes Jev as trained for calibrated decisions through Reinforcement Learning for Calibrated Decisions (RLCD).\footnote{TypeSafe's \href{https://typesafe.ai/blog/introducing-system-one-models-and-jev}{model introduction} and \href{https://docs.typesafe.ai/introduction/machine-learning-primer}{machine-learning primer} describe this training goal; the complete architecture and training recipe remain proprietary.} Recent efforts to reproduce this behavior have produced a growing family of models \citep{inner,llmasjev,startlux}, which we call \emph{Jev reproductions}. These implementations range from training a decision head with backbone adaptation to extracting candidate probabilities from a frozen LLM. The term denotes a shared behavioral goal, without implying recovery of Jev's proprietary internals or a common architecture. Candidate representation, probability readout, and decision-specific learning are therefore possible sources of robustness, not guarantees of it.

Existing evidence already challenges the assumption that structured decisions resist bias. Typed decision models can repeat pairwise judgments reliably yet reverse them when the two responses are swapped \citep{tdmjudges}. In public Jev-style implementations, option labels can override written definitions, and serialization changes can alter this behavior with model weights fixed \citep{labeldefinitions}. What remains limited is a systematic comparison of Jev versions, Jev reproductions, and generative LLMs as correct choices face repeated reordering in multi-option tasks. A single swap cannot reveal how widely failures spread over further orders, and aggregate accuracy can hide losses that recoveries offset. We therefore ask: \emph{does Jev-style decision structure reduce selection bias?}

We investigate this question with a simple test that shuffles options while preserving question and candidate content. The study compares two Jev versions, twelve Jev reproductions, and six generative LLMs on SuperGPQA \citep{supergpqa}, LEXam \citep{lexam}, MMLU-Pro \citep{mmlupro}, and GPQA Diamond \citep{gpqa}. Our primary collection pools 1,700 questions from SuperGPQA and LEXam. For each question, we compare the original order, its reversal, and six random permutations with eight requests in the unchanged original order. Tracking first errors reveals how the failure set expands and how many initial successes survive. Every request retains the full candidate menu.

The test reveals no general Jev advantage in preserving correct choices. Jev's accuracy varies by less than two percentage points, yet its cumulative failure set expands by 27.3--28.5\%, with first failures continuing into the final presentation (Figure~\ref{fig:order-overview}). Four LLMs preserve more initial successes, including on questions every model initially answers correctly. Qwen3.8-Flash-Next and GLM-5.3-Flash have lower pooled retention, preventing a universal ranking of model families. The central finding is that Jev's structured decision formulation does not by itself protect answers the model can already get right.

Fixed-order repetition exposes a second distinction: repeatability does not ensure robustness to reordering. Both Jev versions repeat choices more consistently than all six LLMs, but lose substantially more initial successes under reordering than under repetition at the same call budget. The twelve Jev reproductions make this gap especially clear: 91.9--100\% fixed-order agreement coexists with losses of 23.2--64.2\% of initial successes under reordering (Figure~\ref{fig:open-order-overview}).

The remaining analyses ask where correct choices are most fragile and whether repeated answers can help recover them. Comparisons by task, difficulty, and menu size locate this vulnerability; voting and unanimity test whether the observed variation or agreement offers a remedy. MMLU-Pro and GPQA Diamond then examine the scope of the findings, showing that correct choices remain vulnerable where Jev achieves higher accuracy and with only four candidates. Together, these analyses distinguish average accuracy, fixed-order repeatability, and preservation of correct choices. The supporting survey in Appendices~\ref{sec:introduction}--\ref{sec:inventory} organizes differences in representation, learning, and computation that controlled comparisons could isolate to explain the observed behavior.

\section{Evaluation Setup}
\label{sec:bias-setup}

\subsection{Datasets}

Our primary evaluation uses a fixed collection of 1,700 test questions: 1,200 from SuperGPQA \citep{supergpqa} and 500 from LEXam \citep{lexam}.\footnote{Dataset releases: \href{https://huggingface.co/datasets/m-a-p/SuperGPQA}{SuperGPQA} and \href{https://huggingface.co/datasets/LEXam-Benchmark/LEXam}{LEXam}.} The benchmarks cover graduate-level knowledge and legal examination questions. We weight questions equally, so pooled scores reflect this sample composition rather than an equal-weight average of benchmark scores.

SuperGPQA contributes questions with 4--10 options from its official evaluation release, balanced across easy, middle, and hard difficulty. LEXam contributes 125 test questions at each of 4, 8, 16, and 32 options. This collection supports analyses by difficulty, discipline, and candidate count. Candidate count varies with benchmark and question content, so these comparisons are descriptive.

Two further datasets test the scope of the findings. MMLU-Pro \citep{mmlupro} contributes 300 test questions, with 100 each at eight, nine, and ten options and 21--22 from each of fourteen subjects. GPQA Diamond \citep{gpqa} contributes all 198 questions, each with four options.\footnote{Dataset releases: \href{https://huggingface.co/datasets/TIGER-Lab/MMLU-Pro}{MMLU-Pro} and \href{https://huggingface.co/datasets/Idavidrein/gpqa}{GPQA}, Diamond subset.} We report these datasets separately to examine whether the vulnerability persists at higher Jev accuracy and with short candidate menus. The balanced MMLU-Pro sample does not estimate accuracy over the full benchmark distribution. All question sets remain fixed across models and presentations.

\subsection{Models}

The Jev--LLM comparison includes Jev 1.13 and Jev 1.14, with DeepSeek-V4.1-Flash-1017, Claude Sonnet 5.5, GPT-6.1-sol, Qwen3.8-Flash-Next, Gemini 3.8 Flash, and GLM-5.3-Flash as generative LLM baselines.\footnote{Model documentation: \href{https://docs.typesafe.ai/models}{Jev}, \href{https://api-docs.deepseek.com/quick_start/pricing}{DeepSeek-V4.1-Flash-1017}, \href{https://platform.claude.com/docs/en/models/overview}{Claude Sonnet 5.5}, \href{https://www.alibabacloud.com/help/en/model-studio/models}{Qwen3.8-Flash-Next}, \href{https://ai.google.dev/gemini-api/docs/models}{Gemini 3.8 Flash}, and \href{https://docs.z.ai/guides/llm/glm-5.3}{GLM-5.3-Flash}.} Jev returns a choice through its native decision interface. The LLMs receive a serialized representation of the same question and candidate menu and return a selection. Their training, inference procedures, and interfaces are not matched.

The comparison with Jev reproductions includes AutoTrust-JEV-9B, AutoTrust-JEV-27B, AutoTrust-JEV-35B, AutoTrust-JEV-27B-v1.1, AutoTrust-JEV-Gemma4-26B-A4B, StartLux-Decision-v1-0.8B, StartLux-Decision-v1-2B, StartLux-Decision-v1-4B, StartLux-Decision-v1-9B, StartLux-Decision-v1-27B, StartLux-Decision-v1-35B-A3B \citep{startlux}, and pplx-decider-v1.1-27b.\footnote{Official model cards: \href{https://huggingface.co/autotrust/JEV-27B}{AutoTrust-JEV-27B}, \href{https://huggingface.co/autotrust/JEV-Gemma4-26B-A4B}{AutoTrust-JEV-Gemma4-26B-A4B}, and \href{https://huggingface.co/perplexity-ai/pplx-decider-v1.1-27b}{pplx-decider-v1.1-27b}.} We select the candidate with the highest native probability, without generating an answer or using deliberation. Both Jev versions serve as references, giving twenty unique models across the two comparisons. Every model is evaluated on all four datasets.

\subsection{Eight Presentations of Each Question}

Across all four datasets, each question is presented in its original candidate order, the reversed order, and six random permutations. All models receive the same permutation for a given question. Each candidate's identifier moves with its content, preserving their association, while the question and state remain unchanged. Every request retains the full menu. These are repeated measurements of fixed questions. In the primary collection, the first random permutation differs from the original, and each subsequent random permutation differs from all preceding permutations. MMLU-Pro has eight distinct orders per question, while GPQA Diamond has seven or eight.

The intervention changes the order of candidate entries without independently manipulating identifiers or rewriting answers. The original presentation provides a reference for trajectories; it is not assumed to be the least biased order. Requests are independent: later presentations do not include earlier model answers.

We request low reasoning effort for DeepSeek-V4.1-Flash-1017, GPT-6.1-sol, Qwen3.8-Flash-Next, Gemini 3.8 Flash, and GLM-5.3-Flash, and use adaptive thinking with low effort for Claude Sonnet 5.5. Returned keys are mapped back to the original candidates. Moving a candidate to a different position does not count as a changed decision if the model selects the same candidate.

\subsection{Repeated Requests in the Original Order}
\label{sec:bias-repeat-method}

The original-order repetition study holds candidate content, identifiers, and order fixed across eight independent requests for every model: the original presentation and seven further requests. Following the distinction between repetition stability and position consistency in \citet{judgingjudges}, this condition measures fixed-order repeatability through same-candidate agreement across requests.

For each comparison, we contrast eight original-order presentations with eight reordered presentations: the original, reversal, and six random permutations. The two conditions share the original request and use the same questions. We compare changed-choice rates, loss of initial successes, mean accuracy, and plurality gains. Sensitivity analyses on the primary collection compare the first three original-order requests with the original and every pair of the seven non-original presentations, and remove the shared original request to compare two further repeats with reversal and the first permutation.

Same-candidate agreement requires selecting the same candidate in every request so far. A changed choice involves at least two distinct candidates. The same-wrong rate counts questions receiving the same wrong candidate throughout. Loss of initial successes is the fraction of first-request successes that become incorrect at least once. Pointwise 95\% intervals use 2,000 benchmark-stratified bootstrap resamples of source groups, jointly preserving models and conditions; differences use paired draws.

\subsection{Separating Scores, Correctness, and Error Identity}
\label{sec:bias-definition}

Let $\mathcal Q$ denote the evaluated collection or subgroup. For question $x\in\mathcal Q$, let $\mathcal A_x$ be its candidate set, $y_x$ its correct candidate, and $\widehat y_{x,t}$ the model's answer in presentation $t\in\{1,\ldots,8\}$. Accuracy is the proportion of questions answered correctly in each presentation. For a given model, define
\begin{equation}
 C_t=\{x\in\mathcal Q:\widehat y_{x,t}=y_x\},\qquad
 R_t=\frac{|\bigcap_{s=1}^{t} C_s|}{|C_1|}.
 \label{eq:bias-retention}
\end{equation}
We call $R_t$ \emph{retention}: the fraction of initially correct questions that remain correct in every presentation through $t$. Its complement, $1-R_t$, is the \emph{loss of initial successes}. A question that becomes wrong and later recovers no longer belongs to the retained set. Retention therefore measures continued correctness, rather than overlap with the correct set in the latest presentation alone. Both quantities are undefined when there are no initial successes.

Within the Jev--LLM comparison, we additionally restrict $\mathcal Q$ to questions answered correctly by all eight models in the original presentation. Each model's retention then uses this same denominator. We report paired differences with 2,000 bootstrap resamples of source groups within each benchmark, jointly preserving all models and rounds. This analysis is conditional on the observed shared initial-correct set; it removes differences in which questions enter the comparison, without matching model capability or training.

To measure how many questions have failed, define the cumulative failed set and the number of first failures in presentation $t$ as
\begin{equation}
 F_t=\bigcup_{s=1}^{t}(\mathcal Q\setminus C_s),\qquad
 B_t=|F_t|,\qquad N_t=|F_t\setminus F_{t-1}|,
 \label{eq:bias-first-failures}
\end{equation}
with $F_0=\varnothing$. Each question enters $F_t$ once, at its first incorrect outcome. Later mistakes on that question add nothing to $N_t$, even if the selected wrong candidate changes. Figure~\ref{fig:order-overview}(c) reports $B_t$. The cumulative count is bounded by the number of questions, and $B_t-B_1=|C_1|(1-R_t)$ links additional failures to the loss of initial successes.

Separately, accuracy discards changes in the identity of an incorrect answer. To track wrong-option diversity, we count distinct question--wrong-option pairs through presentation $t$:
\begin{equation}
 U_t=\left|\bigcup_{s=1}^{t}
 \{(x,a):x\in\mathcal Q,\ a\in\mathcal A_x\setminus\{y_x\},\ \widehat y_{x,s}=a\}\right|.
 \label{eq:bias-error-coverage}
\end{equation}
Selecting the same wrong option again on the same question adds nothing to $U_t$; selecting another wrong option adds one. Consequently, $U_t$ may exceed the number of questions ever answered incorrectly. It is nondecreasing by construction. The empirical quantities of interest are how many new pairs each presentation adds and how often a single question receives multiple distinct wrong answers.

To separate question coverage from wrong-option diversity, let $k_{x,t}$ be the number of distinct wrong options selected for question $x$ through presentation $t$, and let $W_t=|\{x\in\mathcal Q:k_{x,t}>0\}|$. Then
\begin{equation}
 U_t = W_t + D_t,\qquad
 D_t = \sum_{x\in\mathcal Q}\max(k_{x,t}-1,0).
 \label{eq:bias-error-decomposition}
\end{equation}
Each question contributes its first observed wrong candidate to $W_t$ and any additional distinct wrong candidates to $D_t$. Since $D_1=0$, growth decomposes as $U_t-U_1=(W_t-W_1)+D_t$.

\subsection{Stratification and Uncertainty}
\label{sec:bias-strata-method}

The pooled collection supports complementary views of stability. Within each full subgroup $\mathcal Q_g$, we report accuracy averaged over the eight presentations, the percentage of questions with at least two distinct selected candidates, loss of initially correct questions ($100(1-R_8)$), and extra wrong pairs per 100 questions ($100D_8/|\mathcal Q_g|$). Changed choices split into questions selecting both the correct candidate and an actual wrong candidate, and questions selecting multiple wrong candidates without ever selecting the correct one. These disjoint groups sum to the changed-choice rate. For the option-count comparison, we also report newly failed questions per 100 questions, $100(B_8-B_1)/|\mathcal Q_g|$. This measures first failures after the original presentation across the full subgroup, whereas $100(1-R_8)$ measures loss conditional on initial success.

We use the source metadata to partition SuperGPQA into easy, middle, and hard questions (400 each) and broad disciplines. We also examine difficulty within the ten-option subset: 264 easy, 258 middle, and 209 hard questions. LEXam groups use option count. Both model comparisons include every observed menu size, from four to 32 options; these groups differ in benchmark composition as well as content. We pool the five SuperGPQA disciplines with fewer than 30 sampled questions for visualization. Difficulty labels are available only for SuperGPQA; we do not assign them to the other benchmarks.

Error bars show pointwise 95\% percentile intervals from 2,000 bootstrap samples. We resample source groups within each benchmark, keeping all presentations and models together within each resampled group. We omit intervals for groups with fewer than 20 questions or eight independent units. These exploratory intervals summarize sampling variation within the observed groups, without simultaneous coverage or a causal interpretation of differences between groups.

\subsection{Voting and Agreement Across Presentations}
\label{sec:bias-consensus-method}

We use the eight selections to examine whether agreement provides useful evidence of correctness. Plurality voting selects the most frequently chosen candidate, without requiring an absolute majority. For ties, we report expected correctness under a uniform draw among the tied candidates. This rule favors neither the original presentation nor the gold answer. We compare voting with mean single-presentation accuracy on $\mathcal Q$. The difference measures the gain from combining eight answers rather than using one of the eight with equal probability.

Unanimous coverage is the fraction of $\mathcal Q$ receiving the same candidate selection in all eight presentations. Accuracy within this subset measures how reliably unanimous answers are correct. We report these quantities together because higher conditional accuracy can come at the cost of answering fewer questions. The pooled bootstrap resamples source groups within each benchmark, preserving all models and presentations together. Voting under original-order repetition also uses eight calls for every model (Section~\ref{sec:bias-repeat-method}).

\section{Does Jev Resist Changes in Option Order?}
\label{sec:bias-results}

The central question concerns the survival of correct choices, not just the stability of a score. We first trace the failures hidden by aggregate accuracy, then compare their extent with LLM baselines and with fixed-order repetition.

\subsection{What Stable Accuracy Conceals}

All eight models in the Jev--LLM comparison maintain similar accuracy across presentations despite substantial differences in their overall performance. Jev scores near 50\%, spanning only 1.47 percentage points for version 1.13 and 1.53 for version 1.14. The LLMs score approximately 56--72\%, with similarly narrow ranges of 1.12--2.59 points (Figure~\ref{fig:order-overview}(a)).

Between two presentations, let $n_{CW}$ count correct-to-incorrect transitions, $n_{WC}$ incorrect-to-correct transitions, and $n_{WW\prime}$ transitions between two distinct wrong candidates. Over $\mathcal Q$, the accuracy change and combined transition rate are
\begin{equation}
 \Delta\mathrm{Acc}_{\mathcal Q}=\frac{n_{WC}-n_{CW}}{|\mathcal Q|},\qquad
 T=\frac{n_{CW}+n_{WC}+n_{WW\prime}}{|\mathcal Q|}.
 \label{eq:bias-cancellation}
\end{equation}
Losses and recoveries offset each other in accuracy but add in the transition rate. Switches between wrong candidates affect only the latter, exposing further variation hidden by a stable score.

\begin{figure}[!htbp]
\centering
\includegraphics[width=\linewidth]{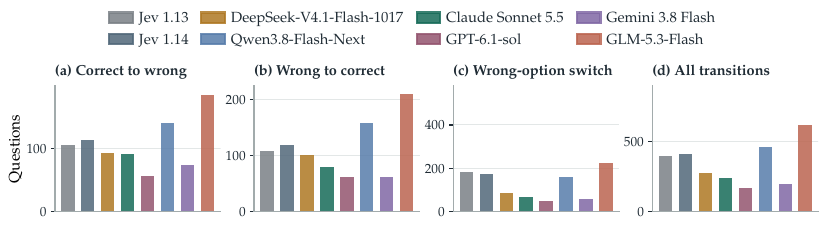}
\caption{\textbf{Accuracy hides changes in individual choices.} Original-to-reversed transitions on all 1,700 questions: (a) correct to incorrect, (b) incorrect to correct, (c) wrong to a different wrong candidate, and (d) all three transition types. Opposing transitions can cancel in accuracy; changes between wrong candidates leave it unchanged.}
\label{fig:paired-transitions}
\end{figure}

\begin{figure}[!htbp]
\centering
\includegraphics[width=\linewidth]{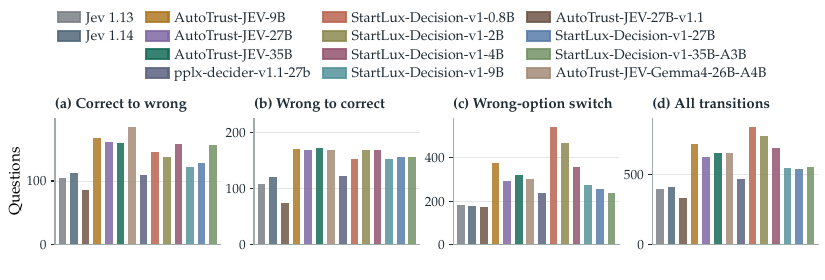}
\caption{\textbf{Accuracy also hides turnover in Jev reproductions.} Reversal transitions on all 1,700 questions, using the categories in Figure~\ref{fig:paired-transitions}. AutoTrust-JEV-9B loses 167 correct answers and recovers 170, while another 377 choices move between wrong candidates.}
\label{fig:open-paired-transitions}
\end{figure}

Jev 1.14 illustrates this cancellation under reversal (Figure~\ref{fig:paired-transitions}): 113 correct answers become incorrect and 120 incorrect answers become correct, yielding a net gain of seven. Another 176 choices move between wrong candidates, bringing the total to 409 transitions. GPT-6.1-sol gains six correct answers despite 170 transitions: 56 losses, 62 recoveries, and 52 wrong-to-wrong switches. Qwen3.8-Flash-Next and GLM-5.3-Flash have 459 and 620 transitions for net gains of only 19 and 26. Similar score changes thus conceal different amounts of turnover, including failures on questions each model could previously answer.

The same cancellation occurs in Jev reproductions (Figure~\ref{fig:open-paired-transitions}). AutoTrust-JEV-9B loses 167 correct answers and recovers 170, while another 377 choices switch between wrong candidates. Its 714 transitions yield a net gain of three correct answers. Across all eight presentations, the twelve Jev reproductions' accuracy ranges are only 1.12--2.18 points (Figure~\ref{fig:open-order-overview}(a)). Stable aggregate performance therefore offers little evidence that individual decisions are preserved.

\subsection{First Failures Continue Beyond the Initial Presentations}

\observation{obs:bias-first-failures}{Jev's stable accuracy hides an expanding set of failed questions.}
Jev 1.13 fails on 845 questions in the original presentation and on 1,076 across all eight, while Jev 1.14 rises from 840 to 1,079 (Figure~\ref{fig:order-overview}(c)). The additional 231 and 239 questions expand the observed failure sets by 27.3\% and 28.5\%, despite the narrow accuracy ranges. The LLMs also expose new failures, ranging from 135 for GPT-6.1-sol to 451 for GLM-5.3-Flash, with DeepSeek-V4.1-Flash-1017, Qwen3.8-Flash-Next, and GLM-5.3-Flash exceeding Jev. Absolute additions alone do not establish a robustness ranking, however, because models begin with different numbers of correct answers available to lose.

First failures continue well beyond the initial reversal. Reversal exposes 105 and 113 for the two Jev versions; the eighth presentation still reveals eleven and eight questions answered correctly in all seven preceding presentations. Cumulative counts are nondecreasing by definition. The empirical finding is the scale of the expansion and the continued loss of questions that have already survived several orders.

All twelve Jev reproductions likewise expose first failures in the eighth presentation, adding 3--14 per model (Figure~\ref{fig:open-order-overview}(c)). Even AutoTrust-JEV-27B-v1.1, which has comparatively favorable retention, expands from 887 failed questions in the original order to 1,076 across the sequence, with 11 first failures in the final presentation.

\subsection{Persistence on Shared Initial Successes}

\observation{find:bias-turnover}{Jev's retention disadvantage persists against four LLMs on shared initial successes.}
To compare the loss of correct choices across models, retention normalizes first-failure expansion by the number of initial successes: $B_8-B_1=|C_1|(1-R_8)$. Across all eight presentations, Jev retains 72.2--73.0\% of its initially correct answers, compared with 75.7--88.9\% for DeepSeek-V4.1-Flash-1017, Claude Sonnet 5.5, GPT-6.1-sol, and Gemini 3.8 Flash, 64.1\% for Qwen3.8-Flash-Next, and 52.0\% for GLM-5.3-Flash (Figure~\ref{fig:order-overview}(d)). Equivalently, 27.0--27.8\% of Jev's initial successes fail at least once. Four LLMs preserve a larger fraction of initial successes than Jev, while the lower retention of Qwen3.8-Flash-Next and GLM-5.3-Flash rules out a blanket Jev--LLM ranking.

This comparison could be affected by which questions each model answers correctly in the first place. We therefore repeat it on the 487 shared initial successes, requiring every model to begin with a correct answer to every included question. Jev 1.13 and Jev 1.14 retain 85.0\% and 84.4\%, respectively. Retention reaches 92.6\% for DeepSeek-V4.1-Flash-1017, 96.1\% for Claude Sonnet 5.5, 98.4\% for GPT-6.1-sol, and 97.7\% for Gemini 3.8 Flash. GPT-6.1-sol has paired advantages of 13.3 percentage points over Jev 1.13 (95\% interval 9.9--16.6) and 14.0 over Jev 1.14 (10.5--17.2). The DeepSeek-V4.1-Flash-1017, Claude Sonnet 5.5, and Gemini 3.8 Flash intervals also exclude zero. Qwen3.8-Flash-Next retains 84.8\%, with differences of $-0.2$ points from Jev 1.13 ($-4.5$--3.7) and 0.4 from Jev 1.14 ($-3.9$--4.5); neither interval establishes a difference. GLM-5.3-Flash retains only 69.6\%, below both Jev versions, with both paired intervals excluding zero.

Matching initial successes therefore preserves Jev's disadvantage relative to DeepSeek-V4.1-Flash-1017, Claude Sonnet 5.5, GPT-6.1-sol, and Gemini 3.8 Flash, while bringing Qwen3.8-Flash-Next close to Jev and leaving GLM-5.3-Flash below it. This comparison controls which questions enter the calculation, without matching model capability, training, or variation under repeated requests.

Retention also varies widely among Jev reproductions. AutoTrust-JEV-27B-v1.1 loses 23.2\% of initial successes, below the Jev reference losses of 27.0\% and 27.8\%. The other eleven reproductions lose 30.0--64.2\%, including 30.0\% for pplx-decider-v1.1-27b and 49.5\% for AutoTrust-JEV-Gemma4-26B-A4B. Native candidate prediction thus supports implementations on both sides of the Jev results. These rates condition on each model's own initial successes, so their spread also reflects differences in the questions each model initially answers correctly.

\subsection{Repeatability Under a Fixed Order}
\label{sec:bias-repeat-results}

\observation{obs:bias-repetition}{Jev's repeatability under a fixed order does not imply robustness to reordering.}
The preceding comparisons leave open how much variation occurs even when the order stays fixed. Eight original-order requests provide this reference (Figure~\ref{fig:order-overview}(b)). Jev 1.13 and Jev 1.14 achieve 84.9\% and 83.6\% same-candidate agreement, exceeding all six LLMs. GPT-6.1-sol reaches 82.0\% and Gemini 3.8 Flash 83.0\%, while GLM-5.3-Flash falls to 42.1\% even without a menu change. The models thus differ substantially in their tendency to change an answer with the menu held fixed.

\begin{figure}[!htbp]
\centering
\includegraphics[width=\linewidth]{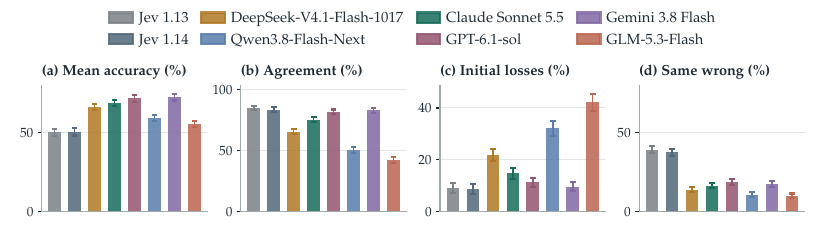}
\caption{\textbf{Jev's fixed-order agreement includes repeated errors.} Eight original-order requests: (a) mean accuracy, (b) agreement across all eight, (c) loss of initial successes, and (d) questions with the same wrong answer throughout. Error bars show pointwise 95\% bootstrap intervals.}
\label{fig:original-consistency}
\end{figure}

High repeatability also includes repeated mistakes (Figure~\ref{fig:original-consistency}). Across eight original-order requests, 37.5--39.3\% of all questions receive the same wrong Jev answer throughout, and accuracy among its unanimous answers is only 53.7--55.1\%. A stable choice therefore provides limited assurance of correctness.

\begin{figure}[!htbp]
\centering
\includegraphics[width=\linewidth]{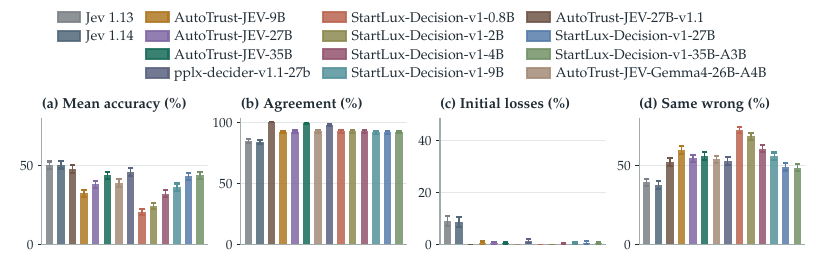}
\caption{\textbf{High fixed-order agreement also repeats errors.} Eight original-order requests, with metrics as in Figure~\ref{fig:original-consistency}. Error bars show pointwise 95\% bootstrap intervals.}
\label{fig:open-original-consistency}
\end{figure}

The Jev reproductions combine even higher agreement with frequent repeated errors (Figure~\ref{fig:open-original-consistency}). They agree on 91.9--100\% of all questions across eight requests, yet 48.5--72.2\% receive the same wrong answer throughout. AutoTrust-JEV-27B-v1.1 repeats the same candidate on every question; pplx-decider-v1.1-27b reaches 97.7\% agreement. Repeatedly selecting the argmax of native candidate probabilities can therefore produce highly stable errors.

\begin{figure}[!htbp]
\centering
\includegraphics[width=\linewidth]{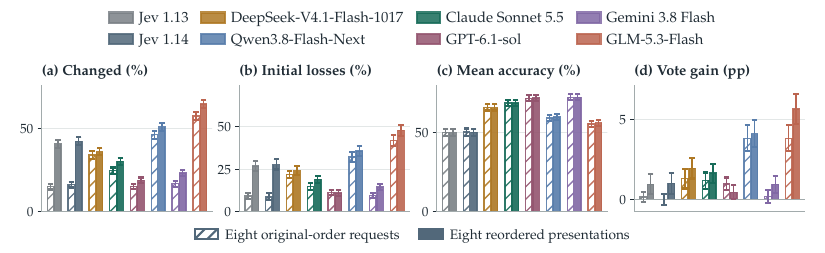}
\caption{\textbf{Reordering exposes more Jev instability at equal call counts.} Eight original-order requests (hatched) versus eight reordered presentations (solid): (a) changed choices, (b) loss of initial successes, (c) mean accuracy, and (d) plurality gain over mean accuracy. Error bars show pointwise 95\% paired bootstrap intervals.}
\label{fig:repeat-vs-reorder}
\end{figure}

At an equal eight-call budget, Jev varies substantially more under reordering than under original-order repetition (Figure~\ref{fig:repeat-vs-reorder}). Its changed-choice rate is 25.8 percentage points higher for version 1.13 (95\% interval 23.7--28.1) and 25.9 higher for version 1.14 (23.6--28.2), compared with differences of 2.1--7.2 across the six LLMs. This gap also extends to successful decisions. Jev loses 8.8--9.2\% of its initial successes under repetition and 27.0--27.8\% under reordering, with paired differences of 17.8 points (15.2--20.5) and 19.0 points (16.1--22.0), while mean accuracy differs by less than half a point. By contrast, GLM-5.3-Flash already loses 42.1\% under original-order repetition, with a further 5.9-point difference under reordering. Variation under repetition and additional variation under reordering are distinct failure patterns.

\begin{figure}[!htbp]
\centering
\includegraphics[width=\linewidth]{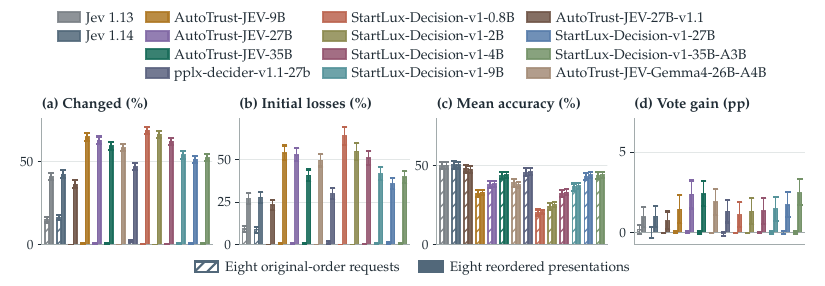}
\caption{\textbf{Reordering reveals failures rarely seen under repetition.} Eight calls per condition, with metrics as in Figure~\ref{fig:repeat-vs-reorder}. Error bars show pointwise 95\% paired bootstrap intervals.}
\label{fig:open-repeat-vs-reorder}
\end{figure}

For the Jev reproductions, the repetition reference makes the contrast sharper: they lose only 0--1.3\% of initial successes under repetition, compared with 23.2--64.2\% under reordering (Figure~\ref{fig:open-repeat-vs-reorder}). Their changed-choice rates are 36.3--68.5 percentage points higher under reordering, versus differences of 25.8 and 25.9 for the Jev references. Choices that are almost perfectly repeatable in a fixed order can thus remain fragile across orders.

Qwen3.8-Flash-Next shows why the overall failure rate and the repetition--reordering gap answer different questions. Its changed-choice rate rises from 46.3\% under repetition to 51.1\% under reordering, an increase of 4.8 points compared with Jev's 25.8--25.9. Its loss of initial successes rises by 3.6 points, from 32.3\% to 35.9\%. Qwen3.8-Flash-Next loses more successes than Jev under reordering, but much of its variability is already present with the menu fixed.

Shorter sequences give the same qualitative result. Combining the original with each of the 21 pairs of non-original presentations yields a three-call comparison against three original-order requests. On the full question set, Jev's excess changed-choice rate is 18.5--21.3 points for version 1.13 and 19.7--21.7 for version 1.14, versus 0.6--6.6 across the LLMs. Removing the shared original request also retains a 12.6--13.5-point excess for Jev. The large repetition--reordering gap is therefore not specific to using all eight presentations or including their shared initial response.

\subsection{Errors Extend Beyond a Single Distractor}

First-failure counts do not distinguish a persistent distractor from several distinct wrong answers on the same question. We therefore examine distinct question--wrong-option pairs, retaining selections after a question's first error.

\observation{find:bias-error-diversity}{Jev's errors often extend beyond one distractor per question.}
If every question with an observed wrong answer had one fixed distractor, wrong-pair growth would reflect only additional questions selecting a wrong candidate. Jev 1.14 illustrates why this description is incomplete. Its original presentation produces 840 wrong pairs, one for each incorrectly answered question. After eight presentations, 1,079 questions have failed at least once, but the model has selected 1,676 distinct wrong pairs. The 836 additional pairs comprise 239 newly affected questions and 597 extra wrong candidates beyond the first per affected question. The latter component includes both initially failed questions and questions that first fail later and then receive another wrong answer.

\begin{figure}[!htbp]
\centering
\includegraphics[width=\linewidth]{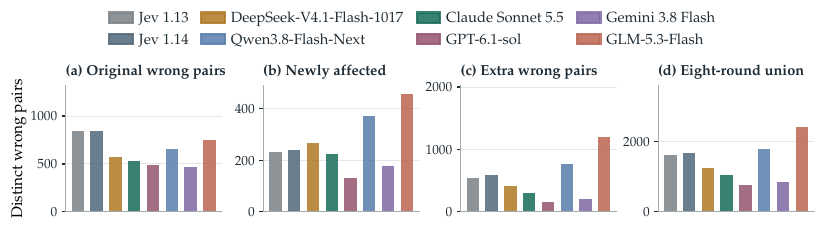}
\caption{\textbf{Additional wrong candidates drive most of Jev's wrong-pair growth.} (a) Original wrong pairs $U_1$, (b) additional questions with an observed wrong candidate $W_8-W_1$, (c) extra wrong candidates beyond the first per affected question $D_8$, and (d) their sum $U_8$. Counts are distinct question--wrong-option pairs and can exceed the number of questions. Every request retains the full menu.}
\label{fig:error-sources}
\end{figure}

Equation~\ref{eq:bias-error-decomposition} separates these two sources for every model (Figure~\ref{fig:error-sources}). Extra wrong candidates account for 70.2--71.4\% of Jev's pair growth, compared with 67.2\% for Qwen3.8-Flash-Next, 72.4\% for GLM-5.3-Flash, and 52.9--60.6\% for the other four LLMs. For example, GPT-6.1-sol's increase from 485 to 771 pairs comprises 133 additional questions with an observed wrong candidate and 153 extra wrong candidates. Most of Jev's pair growth thus reflects additional ways to fail on affected questions, beyond the spread of failures counted in Obs.~\ref{obs:bias-first-failures}.

\begin{figure}[!htbp]
\centering
\includegraphics[width=\linewidth]{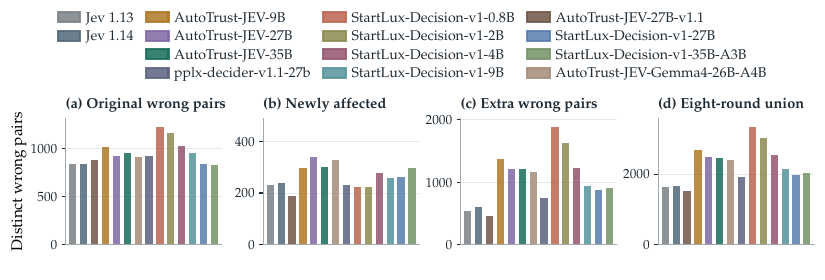}
\caption{\textbf{Additional distractors dominate wrong-pair growth in Jev reproductions.} (a) Original wrong pairs $U_1$, (b) newly observed questions with an actual wrong candidate $W_8-W_1$, (c) extra wrong candidates $D_8$, and (d) all distinct wrong pairs $U_8$. Extra candidates account for 70.8--89.5\% of growth.}
\label{fig:open-error-sources}
\end{figure}

Extra wrong candidates account for 70.8--89.5\% of wrong-pair growth in the twelve Jev reproductions, from AutoTrust-JEV-27B-v1.1 to StartLux-Decision-v1-0.8B (Figure~\ref{fig:open-error-sources}). High fixed-order agreement can thus coexist with several distinct wrong answers to the same question under reordering.

Jev's distinct wrong-pair count increases by 91.8--99.5\%, far more than its 27.3--28.5\% increase in failed questions. This is not the largest relative pair growth: the LLM increases range from 59.0\% for GPT-6.1-sol to 220.1\% for GLM-5.3-Flash, with DeepSeek-V4.1-Flash-1017 and Qwen3.8-Flash-Next also exceeding Jev. The distinction is between spreading failure to more questions and exposing more ways to fail on a question.

Multiple distractors are selected on 394--419 questions for Jev, below Qwen3.8-Flash-Next's 451 and GLM-5.3-Flash's 663, and above the 117--261 for the other four LLMs. Their contribution persists late in the sequence. Of the 49 and 63 new wrong pairs in Jev's final presentation, 38 and 55 select a different wrong candidate on a previously affected question, accounting for 77.6\% and 87.3\% of the additions. Qwen3.8-Flash-Next and DeepSeek-V4.1-Flash-1017 show a similar pattern, with shares of 76.8\% and 78.6\%. Late error discovery therefore mainly broadens the observed ways to fail within affected questions, even as a smaller number of previously successful questions fail for the first time.

The expansion is not specific to choosing the original or final presentation as a reference. Relative to the mean wrong-pair count of a single presentation, the full union is 90.5--97.6\% larger for Jev and 60.4--225.9\% larger for the LLMs. Averaging over all eight ways to omit one presentation leaves 51.6--56.9 unseen pairs for Jev, versus 20.5--147.9 for the LLMs. These comparisons characterize the diversity of observed errors without predicting saturation under further evaluation. Because every request retains the full menu, they also leave open how the model would respond if a wrong candidate were removed.

\section{How Does Jev's Order Robustness Vary Across Tasks?}
\label{sec:bias-strata}

The pooled results establish that Jev's stable score conceals lost successes. We now ask how broadly this pattern holds across benchmarks, difficulty levels, topics, and menu sizes. Each profile separates changed choices from lost initial successes, since a model can appear stable by remaining wrong. These subgroup comparisons locate vulnerability without isolating its causes.

\subsection{Comparisons Across Benchmarks}

\begin{figure}[!htbp]
\centering
\includegraphics[width=\linewidth]{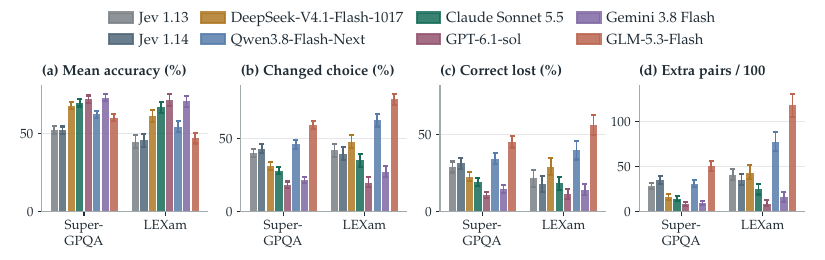}
\caption{\textbf{Jev's relative stability depends on the task and baseline.} By benchmark: (a) eight-presentation mean accuracy, (b) changed choices, (c) loss of initial successes, and (d) extra wrong pairs beyond the first per affected question, per 100 questions. Error bars show pointwise 95\% bootstrap intervals (Section~\ref{sec:bias-strata-method}).}
\label{fig:benchmark-profiles}
\end{figure}

\observation{find:bias-task}{Jev's relative stability depends on the task and the LLM comparator.}
The ordering of models by stability can change across benchmarks (Figure~\ref{fig:benchmark-profiles}). On LEXam, Jev changes 39.8--42.0\% of choices, below DeepSeek-V4.1-Flash-1017's 48.0\%, Qwen3.8-Flash-Next's 62.8\%, and GLM-5.3-Flash's 77.0\%, but above GPT-6.1-sol, Gemini 3.8 Flash, and Claude Sonnet 5.5. DeepSeek-V4.1-Flash-1017 instead changes fewer choices than either Jev version on SuperGPQA. GPT-6.1-sol and Gemini 3.8 Flash remain more stable than Jev on both benchmarks, while Qwen3.8-Flash-Next and GLM-5.3-Flash remain less stable. Jev's relative position therefore depends on both the task and the comparator.

\begin{figure}[!htbp]
\centering
\includegraphics[width=\linewidth]{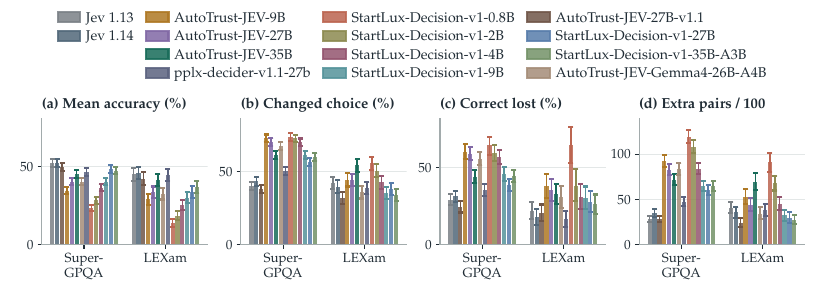}
\caption{\textbf{Lower turnover need not mean better retention.} Benchmark profiles of Jev reproductions, with metrics as in Figure~\ref{fig:benchmark-profiles}. AutoTrust-JEV-27B-v1.1 changes fewer choices than either Jev version on both primary benchmarks, yet loses more initial successes than Jev 1.14 on LEXam. Error bars show pointwise 95\% bootstrap intervals.}
\label{fig:open-benchmark-profiles}
\end{figure}

AutoTrust-JEV-27B-v1.1 changes fewer choices than either Jev version on both primary benchmarks, but this does not translate into uniformly better retention (Figure~\ref{fig:open-benchmark-profiles}). On LEXam, its loss of 20.5\% lies between Jev 1.13's 21.7\% and Jev 1.14's 17.9\%, despite changing fewer choices than both. The distinction between preserving candidate identity and preserving correct answers therefore matters even among Jev-style models.

The changed choices themselves also differ in whether they ever reach a correct answer. Over half of Jev's LEXam questions with changed choices remain wrong in every presentation. These always-wrong switches account for 21.4--21.6\% of all LEXam questions, compared with 5.2--20.6\% for the LLMs. Such trajectories contain no correct answer for voting to recover, linking the stability analysis to the aggregation results in Section~\ref{sec:bias-consensus}.

\subsection{Difficulty and the Persistence of Correct Answers}

\begin{figure}[!htbp]
\centering
\includegraphics[width=\linewidth]{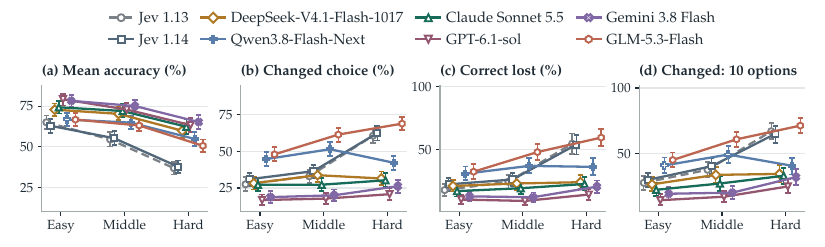}
\caption{\textbf{Jev changes more choices on harder questions.} SuperGPQA difficulty groups: (a) mean accuracy, (b) changed choices, (c) loss of initial successes, and (d) changed choices restricted to ten-option questions. Error bars show pointwise 95\% bootstrap intervals. Panel (d) holds menu size fixed; question content still differs.}
\label{fig:difficulty-profiles}
\end{figure}

\observation{find:bias-difficulty}{Jev loses over half its initial successes on hard SuperGPQA questions.}
Hard questions challenge Jev beyond lowering its accuracy. Even conditional on an initially correct answer, 53.9--54.6\% later become wrong, compared with 14.1--36.2\% for five LLMs and 59.5\% for GLM-5.3-Flash (Figure~\ref{fig:difficulty-profiles}). An initial Jev success on these questions is therefore more often lost than preserved across all eight presentations. Jev's overall changed-choice rate also rises from 27.5--31.0\% on easy questions to 60.8--62.5\% on hard questions. Every LLM shows a smaller increase, and Qwen3.8-Flash-Next's rate decreases. For Jev, harder questions expose both more frequent changes and a greater loss of initially correct answers.

\begin{figure}[!htbp]
\centering
\includegraphics[width=\linewidth]{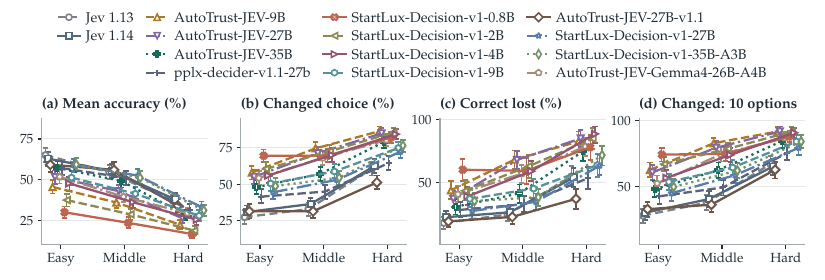}
\caption{\textbf{Hard questions expose greater order sensitivity in Jev reproductions.} SuperGPQA profiles, as in Figure~\ref{fig:difficulty-profiles}. Every reproduction changes more choices on hard than easy questions, including within ten-option menus. Error bars show pointwise 95\% bootstrap intervals; content and subject composition differ.}
\label{fig:open-difficulty-profiles}
\end{figure}

The difficulty gradient extends across the twelve Jev reproductions (Figure~\ref{fig:open-difficulty-profiles}). Every model changes more choices on hard than easy SuperGPQA questions, including within ten-option menus. On hard questions overall, loss of initial successes ranges from 36.8\% for AutoTrust-JEV-27B-v1.1 to 88.5\% for StartLux-Decision-v1-4B. High fixed-order agreement thus does not protect a correct answer once the model confronts harder questions in a different order.

Menu size does not fully explain this gradient. Within ten-option questions, Jev's changed-choice rate rises from 27.7--29.9\% on easy questions to 65.1--67.0\% on hard questions; every LLM shows a smaller increase (Figure~\ref{fig:difficulty-profiles}(d)). Question content and subject composition still differ, so this comparison does not isolate difficulty itself.

\begin{figure}[!htbp]
\centering
\includegraphics[width=\linewidth]{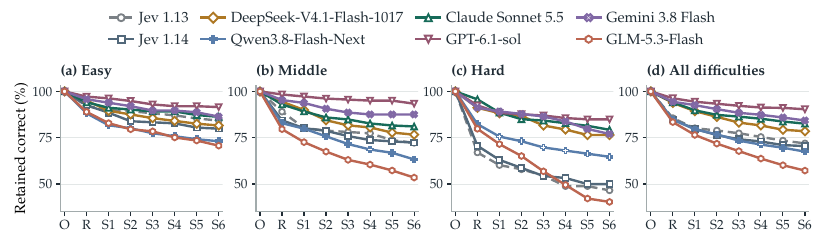}
\caption{\textbf{Harder questions amplify Jev's loss of correct answers.} Retention across eight presentations for ten-option SuperGPQA questions: easy, middle, hard, and all difficulties. Each curve starts at 100\% and retains a question only while every observed answer remains correct. Panels share a scale; question content differs across groups.}
\label{fig:difficulty-retention}
\end{figure}

Correct-answer retention shows that the vulnerability is present even on easy questions (Figure~\ref{fig:difficulty-retention}). Within ten-option groups, Jev loses 15.7--20.1\% of initial successes on easy questions, 27.7\% on middle questions, and 50.0--53.4\% on hard questions. Across all 731 ten-option questions, the loss is 28.0--29.5\%, while each version's accuracy varies by only 2.05 percentage points. Hard questions amplify the discrepancy between stable accuracy and fragile successes; they are not necessary for it to appear.

\begin{figure}[!htbp]
\centering
\includegraphics[width=\linewidth]{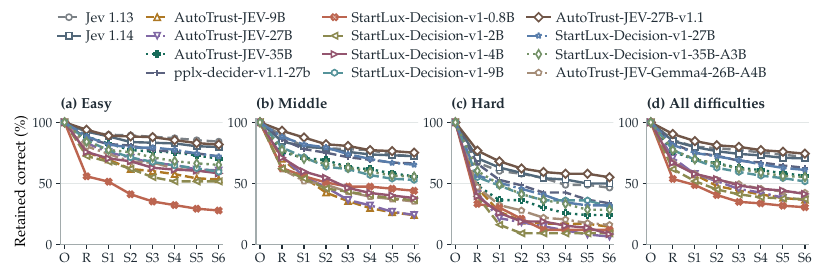}
\caption{\textbf{Correct choices by Jev reproductions are more fragile on hard questions.} Ten-option SuperGPQA retention by difficulty. AutoTrust-JEV-27B-v1.1 loses 18.2\% of initial successes on easy questions and 44.9\% on hard questions. Panels share a scale, whose range differs from Figure~\ref{fig:difficulty-retention}.}
\label{fig:open-difficulty-retention}
\end{figure}

The same ten-option analysis separates the Jev reproductions (Figure~\ref{fig:open-difficulty-retention}). AutoTrust-JEV-27B-v1.1 loses 18.2\% of initial successes on easy questions and 44.9\% on hard questions. The other eleven lose 66.2--93.3\% on hard questions, compared with 50.0--53.4\% for the Jev references in this comparison. A common candidate count therefore leaves substantial differences both across difficulty groups and within the Jev-style family.

\begin{figure}[!htbp]
\centering
\includegraphics[width=\linewidth]{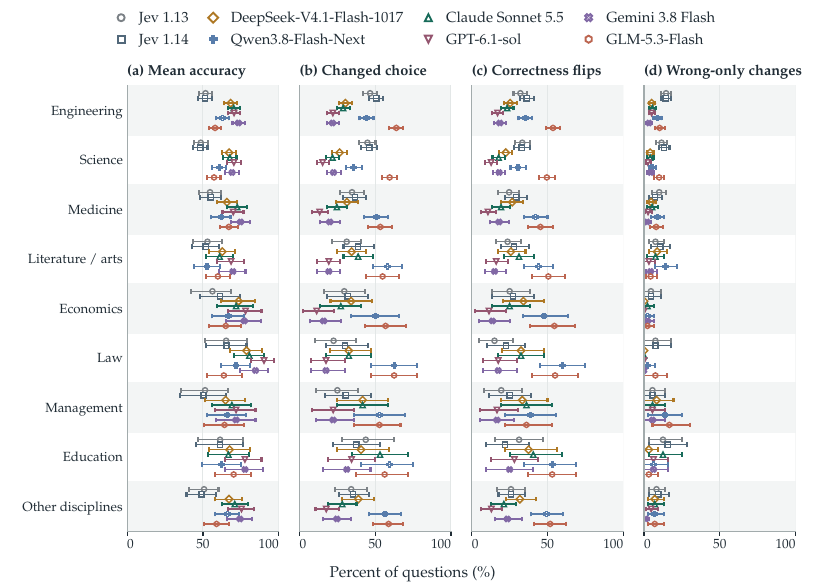}
\caption{\textbf{Subject profiles separate switches involving correct and wrong candidates.} By SuperGPQA discipline: (a) mean accuracy, (b) changed choices, (c) both correct and actual wrong candidates, and (d) multiple wrong candidates without a correct answer. Rates in (c,d) sum to (b). Error bars show pointwise 95\% bootstrap intervals. Other disciplines pools five groups with fewer than 30 questions each; subject and difficulty vary together.}
\label{fig:supergpqa-topics}
\end{figure}

Subject profiles help locate this difficulty gradient (Figure~\ref{fig:supergpqa-topics}). Engineering and science contain much of Jev's instability and many of the hard questions: 163 of 399 engineering questions and 142 of 321 science questions are hard, compared with nine of 93 in literature and arts. Their higher changed-choice rates are consistent with the difficulty result, without identifying an independent subject effect. Separating switches that include the correct candidate from always-wrong switches also reveals which variation includes an answer that aggregation could recover.

\begin{figure}[!htbp]
\centering
\includegraphics[width=\linewidth]{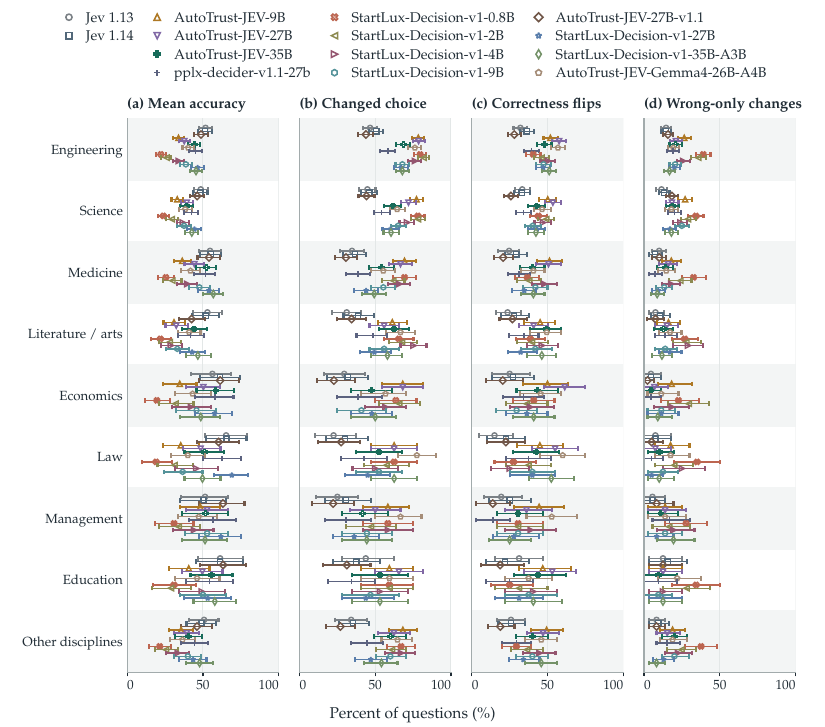}
\caption{\textbf{Jev reproductions switch between wrong answers within subjects.} SuperGPQA discipline profiles, using the decomposition in Figure~\ref{fig:supergpqa-topics}. Switches involving the correct candidate and always-wrong switches partition each model's changed-choice rate. Error bars show pointwise 95\% bootstrap intervals; subject and difficulty vary together.}
\label{fig:open-supergpqa-topics}
\end{figure}

Subject profiles also qualify the relatively favorable pooled result for AutoTrust-JEV-27B-v1.1 (Figure~\ref{fig:open-supergpqa-topics}). It changes 43.9\% of choices in engineering and 44.2\% in science. Always-wrong switches account for 15.8\% and 18.4\% of all questions in those disciplines, respectively. Thus, its residual instability includes both lost or recovered correct answers and variation that never reaches the correct answer; the latter cannot be repaired by voting over the recorded choices.

\subsection{Locating Instability Within Legal Tasks}

\begin{figure}[!htbp]
\centering
\includegraphics[width=\linewidth]{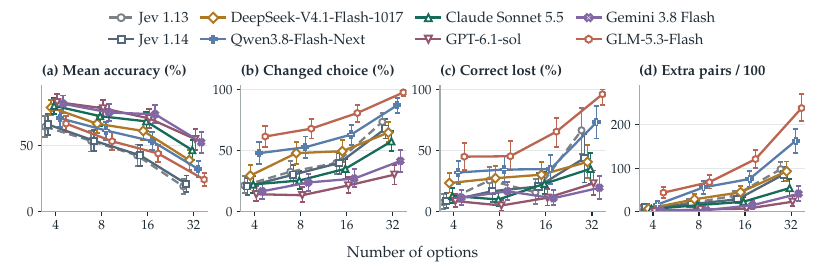}
\caption{\textbf{Larger LEXam menus accompany more varied wrong choices.} Option-count groups contain 125 questions each; metrics follow Figure~\ref{fig:benchmark-profiles}. Lines connect model estimates, with pointwise 95\% bootstrap intervals. Groups contain different questions, so the comparison does not isolate an effect of option count.}
\label{fig:lexam-options}
\end{figure}

LEXam reveals substantial heterogeneity across menu sizes. All eight models in the Jev--LLM comparison have lower accuracy and more changed choices in the 32-option group than in the four-option group (Figure~\ref{fig:lexam-options}). Jev's changed-choice rates rise from 20.8--21.6\% in the four-option group to 68.0--73.6\% in the 32-option group. Both Jev versions exceed DeepSeek-V4.1-Flash-1017's 64.8\% in the latter group, reversing their favorable comparison on LEXam overall. Qwen3.8-Flash-Next and GLM-5.3-Flash vary even more, changing 87.2\% and 97.6\% of choices. The same group yields 85.6--97.6 extra wrong pairs per 100 questions for Jev, compared with 162.4 for Qwen3.8-Flash-Next and 237.6 for GLM-5.3-Flash. Larger menus thus identify a particularly unstable portion of this collection, with wrong-option diversity extending well beyond Jev. The groups contain different questions, however, and do not isolate the effect of adding distractors to a fixed question.

\begin{figure}[!htbp]
\centering
\includegraphics[width=\linewidth]{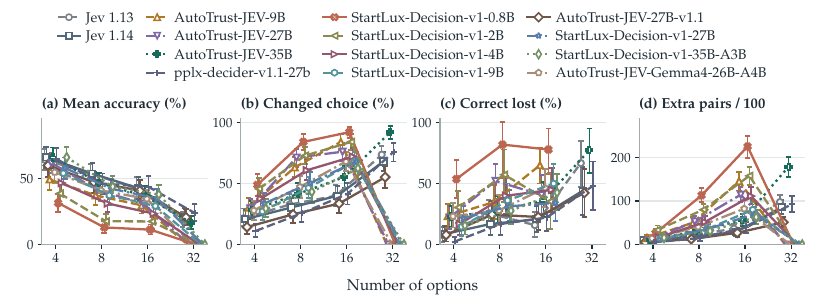}
\caption{\textbf{Jev reproductions vary across LEXam menu sizes.} Profiles include 4-, 8-, 16-, and 32-option questions, using the metrics in Figure~\ref{fig:lexam-options}. Error bars show pointwise 95\% bootstrap intervals. Groups contain different questions.}
\label{fig:open-lexam-options}
\end{figure}

Every reproduction changes more choices at 16 options than at eight, and more at eight than at four (Figure~\ref{fig:open-lexam-options}). AutoTrust-JEV-27B-v1.1 rises from 14.4\% to 33.6\% between four and 16 options, then reaches 55.2\% at 32. Across four to 32 options, pplx-decider-v1.1-27b rises from 11.2\% to 76.0\%; AutoTrust-JEV-35B reaches 92.0\% in the 32-option group.

\subsection{Option Count and Continued Failure Discovery}
\label{sec:bias-option-count}

\begin{figure}[!htbp]
\centering
\includegraphics[width=\linewidth]{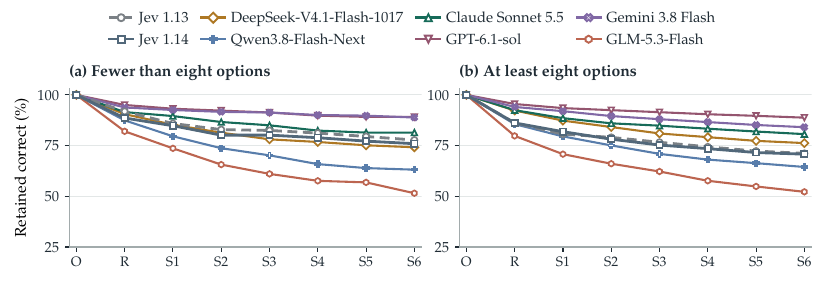}
\caption{\textbf{Jev loses initial successes in both option-count groups.} Retention through each presentation for (a) 437 questions with fewer than eight options and (b) 1,263 with at least eight. Each curve is normalized by the model's initially correct answers in its group.}
\label{fig:option-count-split}
\end{figure}

The LEXam profiles raise a broader question: is the loss of correct choices concentrated in long menus? Across the primary collection, the loss persists on both sides of an eight-option threshold (Figure~\ref{fig:option-count-split}). Jev loses 22.2--24.1\% of its initially correct answers on questions with fewer than eight options and 28.8--29.2\% on questions with at least eight. The comparison with LLMs also varies across these groups. DeepSeek-V4.1-Flash-1017 loses 25.8\% and 23.8\%, respectively: slightly more than either Jev version in the smaller-menu group, but less than both in the larger-menu group. Qwen3.8-Flash-Next loses 36.9\% and 35.5\%, exceeding both Jev versions in each group. These differences qualify pooled comparisons without isolating an effect of menu size.

\begin{figure}[!htbp]
\centering
\includegraphics[width=\linewidth]{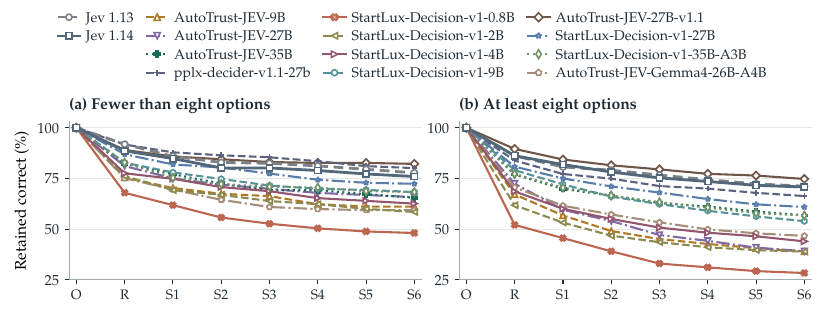}
\caption{\textbf{Jev reproductions lose correct choices even with at least eight options.} Groups and metric follow Figure~\ref{fig:option-count-split}.}
\label{fig:open-option-count-split}
\end{figure}

Jev reproductions also lose initial successes in both groups (Figure~\ref{fig:open-option-count-split}). Losses range from 17.8--51.9\% below eight options and 25.3--71.6\% at eight or more. AutoTrust-JEV-27B-v1.1 has lower loss than either Jev version in both groups. The other eleven reproductions lose a larger fraction of their initial successes than either Jev version in the larger-menu group, showing that the vulnerability extends across the evaluated implementations.

\begin{figure}[!htbp]
\centering
\includegraphics[width=\linewidth]{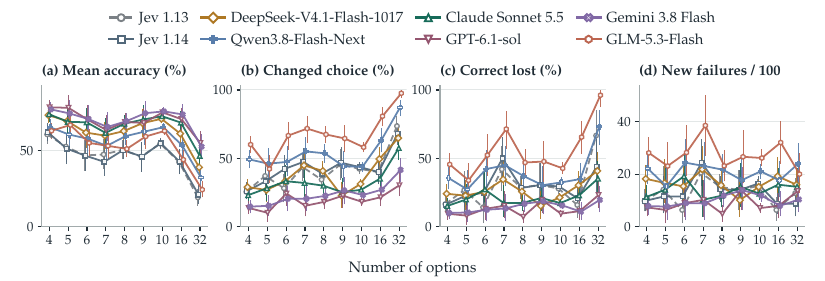}
\caption{\textbf{Jev's instability has no uniform trend across menu sizes.} All observed option counts: (a) mean accuracy, (b) changed choices, (c) loss conditional on initial success, and (d) newly failed questions after O per 100 questions, counted once. Error bars show pointwise 95\% bootstrap intervals. Benchmark and difficulty composition differ across groups.}
\label{fig:all-option-counts}
\end{figure}

The full option-count profile qualifies the LEXam trend (Figure~\ref{fig:all-option-counts}). Jev's changed-choice rate and loss of initial successes fluctuate across the four- to ten-option groups rather than increasing steadily with menu size. Composition matters: all ten-option questions are from SuperGPQA, while all 16- and 32-option questions are from LEXam. The profile identifies vulnerable groups without estimating what would happen if distractors were added to the same question.

\begin{figure}[!htbp]
\centering
\includegraphics[width=\linewidth]{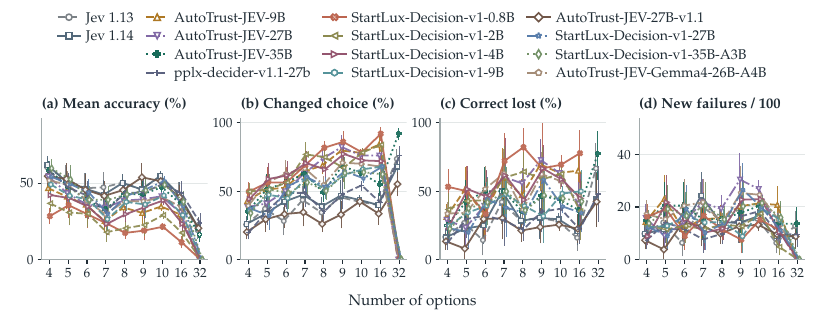}
\caption{\textbf{Jev reproductions fail across the full range of menu sizes.} Four to 32 options; metrics follow Figure~\ref{fig:all-option-counts}. Error bars show pointwise 95\% bootstrap intervals. Benchmark and question content differ across groups.}
\label{fig:open-all-option-counts}
\end{figure}

The pooled profile of Jev reproductions similarly prevents a simple ranking by menu size (Figure~\ref{fig:open-all-option-counts}). AutoTrust-JEV-27B-v1.1 loses 25.6\% of initial successes on ten-option questions and 22.6\% on 16-option questions, although its LEXam-only turnover increases across menu sizes. These groups differ in both benchmark and question content.

Small menus do not make repeated reordering uninformative. A four-option question has $4!=24$ full permutations, so eight presentations need not repeat the same menu order. In the four-option group, 199 of 203 questions have eight distinct observed orders and the other four have seven. Jev 1.13's cumulative failed-question count rises from 83 to 103, and Jev 1.14's from 80 to 100. Presentations five through eight still expose five and two first failures, respectively. Counting only the first occurrence of each full permutation leaves both Jev failure trajectories unchanged. These distinct full orders need not place the correct answer in distinct positions. The curves measure robustness across the sampled menu orders, without assuming a new correct-answer position on every presentation.

The number of new failures must also be read alongside the pool of initial successes. On 32-option questions, Jev loses 44.0--66.7\% of initial successes, but this amounts to only 8.8--12.8 newly failed questions per 100. The ten-option rates are 28.0--29.5\% and 15.5--16.4 per 100. Low initial accuracy leaves fewer successes available to fail, so a smaller cumulative increase can accompany greater conditional fragility. Both measures are needed to avoid mistaking a small remaining pool of correct answers for robustness.

\section{Can Aggregating Reordered Choices Correct Jev's Errors?}
\label{sec:bias-consensus}

\begin{figure}[!htbp]
\centering
\includegraphics[width=\linewidth]{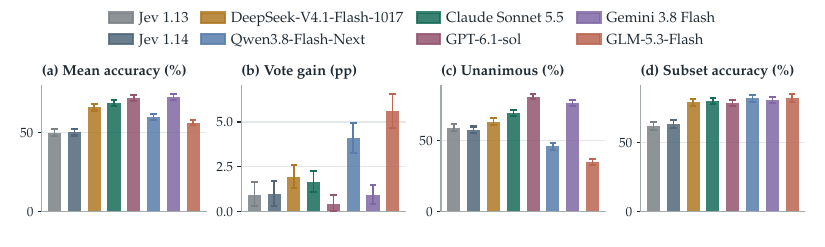}
\caption{\textbf{Voting offers limited Jev correction; unanimity still includes errors.} Across eight orders: (a) mean accuracy, (b) plurality gain in percentage points, (c) unanimous coverage, and (d) accuracy among unanimous answers. Error bars show pointwise 95\% paired source-group bootstrap intervals, stratified by benchmark. Plurality ties use uniform tie-breaking in expectation.}
\label{fig:consensus-profiles}
\end{figure}

The same repeated answers that expose failures might also help correct them. Plurality voting could recover a correct answer that appears across several orders, while unanimity could identify a more reliable subset. We test both possibilities, asking whether variation supplies useful correction and whether agreement warrants confidence (Figure~\ref{fig:consensus-profiles}).

\begin{figure}[!htbp]
\centering
\includegraphics[width=\linewidth]{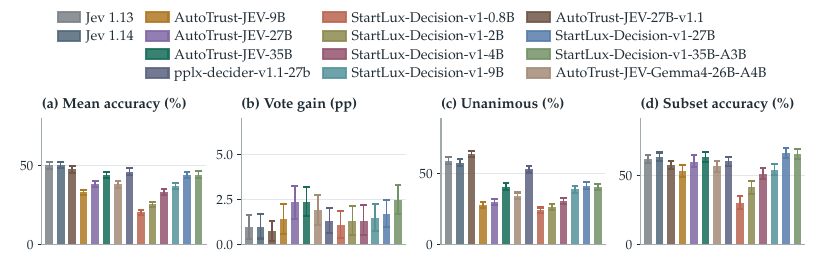}
\caption{\textbf{Voting and unanimity provide limited safeguards for Jev reproductions.} Metrics follow Figure~\ref{fig:consensus-profiles}. Unanimous accuracy ranges from 30.2\% to 66.3\% at differing coverage. Error bars show pointwise 95\% paired bootstrap intervals; plurality ties use uniform tie-breaking in expectation.}
\label{fig:open-consensus-profiles}
\end{figure}

\subsection{Voting Recovers Only Part of the Error}

Combining eight reordered answers gives Jev only a modest accuracy gain: 0.96 percentage points for version 1.13 (95\% interval 0.32--1.64) and 1.01 for version 1.14 (0.34--1.70). The reference is mean single-presentation accuracy, avoiding dependence on an unusually weak original round. LLM gains range from 0.46 points for GPT-6.1-sol to 5.63 for GLM-5.3-Flash, with positive lower bounds for all six intervals. Qwen3.8-Flash-Next and GLM-5.3-Flash gain 4.11 points (3.26--4.94) and 5.63 (4.67--6.55), respectively. Their lower pooled retention thus coexists with more effective correction through voting.

The trajectories help explain why correction is limited. Jev 1.14 gains only 0.30 points on LEXam ($-0.65$--1.28), where many changed choices remain wrong throughout. Such trajectories supply no correct vote; even when the correct answer appears, it must receive enough votes to defeat the alternatives. Qwen3.8-Flash-Next and GLM-5.3-Flash gain 3.32 points (1.87--4.86) and 4.73 (2.99--6.67) on LEXam, while the GPT-6.1-sol and Gemini 3.8 Flash intervals span zero. The usefulness of variation depends on which candidates are selected and how often, rather than on the amount of disagreement alone.

The eight-call comparison further limits the practical benefit of Jev's variation. Voting gain, measured relative to each condition's mean single-request accuracy, is only 0.79 points higher under reordering for Jev 1.13 (95\% interval 0.10--1.47) and 1.00 higher for Jev 1.14 (0.24--1.76), alongside a roughly 26-point difference in changed-choice rates. The direct differences in plurality accuracy between conditions have intervals spanning zero for both versions. Thus, the greater variation under reordering accompanies a small additional voting gain relative to a single answer, without establishing higher final voting accuracy than original-order repetition.

Voting gains for Jev reproductions are also modest despite substantial turnover (Figure~\ref{fig:open-consensus-profiles}). They range from 0.74 points for AutoTrust-JEV-27B-v1.1 (95\% interval 0.18--1.31) to 2.49 for StartLux-Decision-v1-35B-A3B (1.71--3.32), with positive lower bounds for all twelve models. Reordering provides some recoverable correct choices, but the gains remain small relative to the widespread changes it exposes.

\subsection{Agreement and Persistent Error}

\observation{find:bias-consensus}{More than one third of unanimous Jev answers are wrong.}
Unanimity raises Jev's conditional accuracy but leaves substantial error among the retained answers. The two versions select the same candidate in all eight presentations on 57.7--59.1\% of questions, with accuracy of 62.1--63.3\% on this subset. Thus, 36.7--37.9\% of unanimous answers are wrong. These persistent mistakes account for 21.2--22.4\% of all questions, each receiving the same wrong Jev answer eight times. They survive even the strongest agreement requirement available in the recorded trajectories. Such mistakes are not, by themselves, evidence of order bias; they show why surviving the reordering test is insufficient to certify a correct decision.

The same unanimity rule retains more accurate answers for each LLM, although at different coverage. The retained LLM answers are 17.8--21.5\% wrong, compared with Jev's 36.7--37.9\%. GPT-6.1-sol has the broadest coverage at 80.9\%, with 78.5\% conditional accuracy, while DeepSeek-V4.1-Flash-1017, Claude Sonnet 5.5, and Gemini 3.8 Flash retain 63.5--76.5\% of questions at 79.1--81.0\% accuracy. Qwen3.8-Flash-Next and GLM-5.3-Flash retain fewer questions than Jev, at 46.0\% and 35.0\% coverage, but reach 81.8\% and 82.2\% accuracy. Agreement improves conditional accuracy within each model, yet the surviving Jev decisions remain substantially less reliable at these observed operating points. Differences in base accuracy and coverage prevent this comparison from isolating the predictive value of agreement or establishing a ranking at matched coverage.

The Jev reproductions reinforce this limitation of unanimity. Their unanimous subsets reach only 30.2--66.3\% accuracy, with coverage ranging from 24.1\% to 63.7\% of the full collection. AutoTrust-JEV-27B-v1.1 has the highest coverage, yet 42.4\% of its unanimous answers are wrong; these persistent errors account for 27.0\% of all 1,700 questions. Across the twelve models, 12.1--27.0\% of questions receive the same wrong answer in every presentation. Even agreement across reordered menus therefore leaves a substantial set of errors undetected.

\FloatBarrier
\section{Does Higher Accuracy Protect Correct Choices?}
\label{sec:bias-mmlupro}

The primary collection leaves open whether fragile successes mainly accompany Jev's relatively low accuracy on those tasks. MMLU-Pro provides a setting where Jev answers most questions correctly, allowing us to examine whether those successes also survive reordering.

\begin{figure}[!htbp]
\centering
\includegraphics[width=\linewidth]{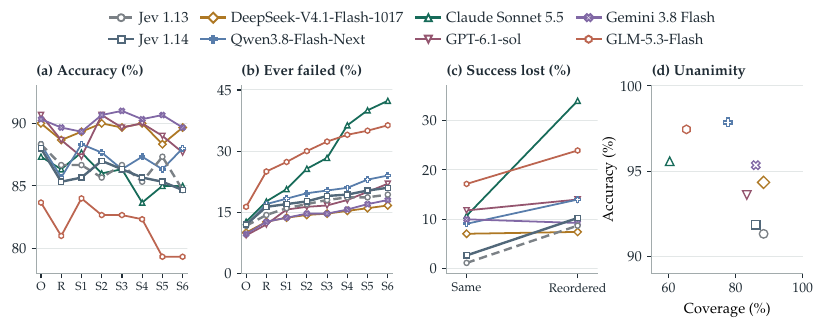}
\caption{\textbf{Jev's higher MMLU-Pro accuracy still leaves correct answers vulnerable.} (a) Accuracy on 300 questions; (b) cumulative failed questions; (c) loss of initial successes after eight original-order or reordered requests; (d) coverage and accuracy of answers unanimous across reordered presentations.}
\label{fig:mmlu-pro}
\end{figure}

\begin{figure}[!htbp]
\centering
\includegraphics[width=\linewidth]{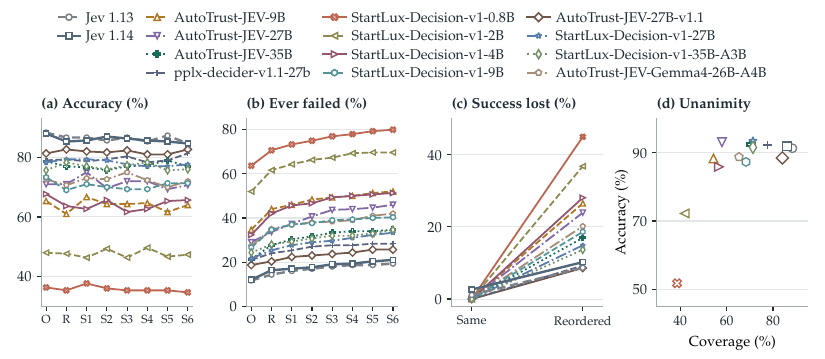}
\caption{\textbf{Perfect repetition does not protect Jev reproductions from reordering.} MMLU-Pro results for twelve Jev reproductions with both Jev versions as references, using the panels in Figure~\ref{fig:mmlu-pro}.}
\label{fig:open-mmlu-pro}
\end{figure}

\observation{obs:bias-mmlupro}{Higher Jev accuracy does not eliminate the loss of initial successes.}
Jev 1.13 and Jev 1.14 average 86.4\% and 86.0\% accuracy over the reordered presentations, with ranges of 3.7 and 3.3 percentage points (Figure~\ref{fig:mmlu-pro}(a)). Yet their cumulative failure sets grow from 35 to 58 and from 36 to 63 questions, respectively. The additional 23 and 27 failures represent 8.7\% and 10.2\% of initially correct answers, compared with losses of only 1.1\% and 2.7\% under original-order repetition. Each version first fails on two more questions in the final presentation. The loss is smaller than on the primary collection, but it persists even when Jev answers most questions correctly.

All twelve Jev reproductions repeat exactly the same candidate across eight original-order requests, yet lose 8.6--45.0\% of initial successes under reordering (Figure~\ref{fig:open-mmlu-pro}). Thus, the separation between repeatability and preservation of correct choices persists on a different benchmark. As in the primary collection, the comparison describes the evaluated systems without isolating an architectural cause.

Jev's unanimous answers are more accurate here than on the primary collection, but agreement still does not certify correctness. They cover 88.3\% and 86.0\% of questions and reach 91.3\% and 91.9\% accuracy (Figure~\ref{fig:mmlu-pro}(d)); 8.7\% and 8.1\% of the retained answers are still wrong. Thus, both the loss under reordering and the reliability of agreement depend on the task collection. What transfers is the distinction among average accuracy, fixed-order repeatability, and the survival of individual correct decisions.

\section{Does a Four-Option Menu Remove the Vulnerability?}
\label{sec:bias-gpqa}

\begin{figure}[!htbp]
\centering
\includegraphics[width=\linewidth]{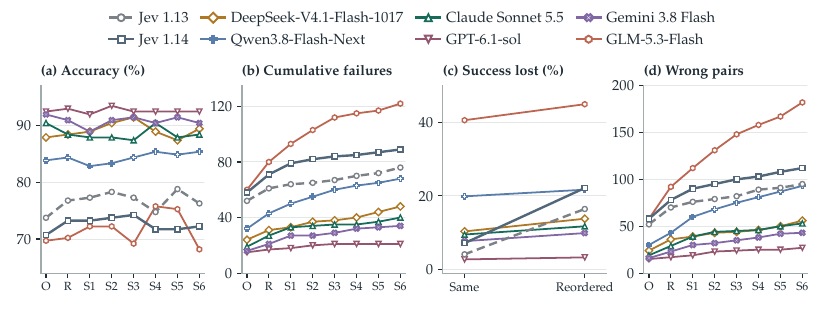}
\caption{\textbf{Four candidates still leave correct Jev choices vulnerable.} GPQA Diamond: (a) accuracy, (b) cumulative failed questions, (c) initial-success loss under repetition and reordering, and (d) distinct wrong question--candidate pairs.}
\label{fig:gpqa-diamond}
\end{figure}

\begin{figure}[!htbp]
\centering
\includegraphics[width=\linewidth]{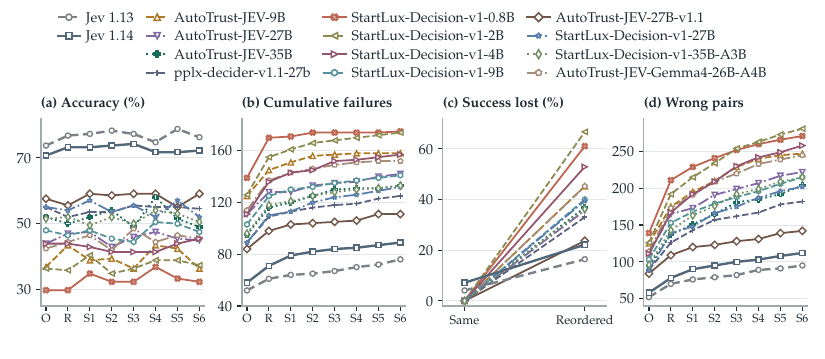}
\caption{\textbf{Perfect repetition coexists with lost successes on four-option questions.} GPQA Diamond results for Jev reproductions, with both Jev versions as references. Panels follow Figure~\ref{fig:gpqa-diamond}.}
\label{fig:open-gpqa-diamond}
\end{figure}

MMLU-Pro extends the evidence to higher Jev accuracy, but its questions still have eight to ten options. GPQA Diamond tests whether the loss of initial successes persists on a separate benchmark where every question has only four options. Figures~\ref{fig:gpqa-diamond} and~\ref{fig:open-gpqa-diamond} use all 198 questions and the common eight-presentation protocol.

\observation{obs:bias-gpqa}{Jev's loss of correct choices is not confined to long candidate menus.}
Jev 1.13 and Jev 1.14 average 76.6\% and 72.6\% accuracy, yet their cumulative failure sets grow from 52 to 76 and from 58 to 89 questions. Reordering loses 16.4\% and 22.1\% of their initial successes, compared with 4.1\% and 7.1\% under original-order repetition. All twelve Jev reproductions repeat their original choices perfectly but lose 23.7--66.7\% of initial successes after reordering. The gap between repeatability and preservation of successful decisions therefore extends to four-option questions. It does not require the larger menus in the primary collection.

The observed loss of initial successes varies across the LLMs, from 3.3\% for GPT-6.1-sol to 21.7\% for Qwen3.8-Flash-Next and 44.9\% for GLM-5.3-Flash. These results do not establish a uniform family ranking. Repeated errors also need not select the same distractor. Jev accumulates 95 and 112 distinct wrong question--candidate pairs across its 76 and 89 failed questions. The reproductions accumulate 142--281 such pairs. Excluding the two questions with duplicate distractor text preserves the Jev repetition--reordering gap. Short menus therefore remain informative about the survival of correct choices, although comparisons across benchmarks do not isolate candidate count from content or difficulty.

\FloatBarrier
\section{Implications and Scope}
\label{sec:bias-discussion}

\paragraph{What the Jev comparison establishes.}
Our results provide no evidence of a general Jev advantage in preserving correct choices across option orders. Four LLMs retain more shared initial successes (Obs.~\ref{find:bias-turnover}), but Qwen3.8-Flash-Next, GLM-5.3-Flash, and the task-dependent comparisons rule out a universal ranking. The contribution beyond pairwise order-sensitivity studies \citep{tdmjudges} is to trace the cumulative consequences: which initial successes are lost, how widely failures spread, and whether further orders still reveal first failures. The magnitude and timing of that loss carry the evidence; monotonic growth of a cumulative set does not.

The Jev reproductions show that a shared decision interface can coexist with substantially different robustness. On the primary collection, AutoTrust-JEV-27B-v1.1 retains more initial successes than the Jev references, whereas the other eleven reproductions retain fewer. Their high fixed-order agreement does not predict this ordering. Explaining the differences requires isolating the choices in representation, learning, and computation that the shared interface leaves unspecified.

\paragraph{Reordering diagnoses failures that agreement may not repair.}
A useful diagnostic need not be an effective correction. Reordering exposes lost successes and multiple wrong candidates, but voting yields only a small accuracy gain for Jev. Always-wrong trajectories contain no correct vote, and unanimous errors pass an agreement filter unchanged (Obs.~\ref{find:bias-consensus}). These persistent errors do not themselves demonstrate order bias; they limit agreement as a safeguard. Nor does observing several wrong candidates establish what would happen after one is removed. An adaptive masking study could compare removing the previously selected wrong candidate with removing a random distractor at matched menu sizes, separating targeted correction from the benefit of fewer choices.

\paragraph{Behavioral differences do not identify their mechanism.}
Original-order repetition characterizes variation without changing the menu. Its comparison with reordering describes behavioral differences without isolating a causal order effect. Candidate identifiers travel with their content while entry order changes, leaving the source of presentation sensitivity unresolved. Prior work distinguishes identifier preferences from position and context effects \citep{labeldefinitions,mcqselectors}; our experiment neither separates these components nor establishes how a proprietary interface represents them.

Training, inference, and interface also differ across models, precluding a causal claim about decision heads or generation. The literature offers starting points for controlled comparisons: frozen LLM readouts keep the backbone fixed \citep{llmasjev}, and serialization interventions keep weights fixed \citep{labeldefinitions}. Further comparisons could vary decision-specific adaptation or candidate interaction separately. The supporting survey organizes these distinctions alongside supervision, gradient paths, and execution. The present findings concern observed selections, not a recovered account of Jev's proprietary mechanism.

\paragraph{The task mixture limits generalization.}
The results cover fixed task collections and eight presentations. Domain, language, difficulty, and candidate count vary together, while subgroup intervals are exploratory and do not adjust for multiple comparisons. MMLU-Pro and GPQA Diamond broaden the evidence, but differences in loss magnitude show its dependence on the collection. Continued first failures establish persistence within the observed sequence without predicting growth under further requests.

\section{Related Work}
\label{sec:bias-related}

\paragraph{Selection bias and its sources.}
Early work on contextual calibration shows that language models carry answer preferences that can destabilize few-shot predictions as prompts and demonstrations change \citep{calibratebeforeuse}. Option reordering provides a more direct test when question and answer content are held fixed \citep{optionsorder}. Position and identifier effects must nevertheless be distinguished. \citet{mcqselectors} identify preferences for option-ID tokens as a major source of multiple-choice selection bias, while FairEval and MT-Bench document response-order effects in pairwise judgments \citep{fairevaluators,mtbench}. In their study of many-shot evaluation, \citet{manyshotevaluators} likewise find that changing response order or the symbols assigned to responses alters GPT-4o's pairwise judgments; they combine evaluations across both transformations to mitigate these biases. \citet{judgingjudges} further separate repetition stability, position consistency, and preference fairness. Our candidate identifiers move with their content, so the present experiment measures behavior across reordered menus without independently identifying these sources.

\paragraph{Mitigation through calibration, training, and reasoning.}
Contextual calibration and PriDe estimate answer priors to reduce bias \citep{calibratebeforeuse,mcqselectors}. CalibraEval extends label-free calibration to LLM judges through an order-preserving transformation of prediction distributions \citep{calibraeval}. Training offers another route. Strengthened symbol binding improves the association between an option and its identifier \citep{symbolbinding}, while permutation-aware GRPO rewards consistency across rearrangements of the same instance \citep{pagrpo}. Recent work also combines a probability-based permutation metric with LoRA fine-tuning and efficient aggregation \citep{permutationlora}. These interventions show that selection bias is responsive to both training and inference procedures. Evidence of reduced position bias in reasoning models on multi-document question answering further motivates comparison with current LLMs \citep{positioncomprehensive}. We evaluate the deployed systems without applying an additional debiasing method to any of them.

\paragraph{Jev-style modeling and selection bias.}
Methods for reproducing Jev-style decisions differ in candidate representation, readout, and learning. LLM-as-Jev can even extract candidate probabilities from frozen causal LLMs \citep{llmasjev}, so the decision formulation alone does not identify an architectural or training intervention. \citet{tdmjudges} compare eight typed decision models with seven LLM judges in pairwise evaluations, finding sensitivity to swapping two responses despite strong repeatability. \citet{labeldefinitions} trace option-label effects to serialization in several public Jev-style implementations through interventions with weights fixed. These studies already establish that typed decisions need not be unbiased. We examine the cumulative consequences in multiple-choice tasks with four to 32 candidates. Repeated permutations reveal how widely failures spread, how much initial success survives, and whether voting or agreement can recover reliable decisions. Fixed-order repetition and contemporary LLM baselines place those losses in context.

\paragraph{Consistency, correctness, and aggregation.}
MT-Bench cautions that a judge can be consistently wrong \citep{mtbench}. GRP evaluates each response pair in both orders and counts it as correct only when both judgments match the gold preference \citep{grp}. Asking for the worse response improves this joint criterion, without isolating gains in order agreement from gains in per-trial correctness. \citet{twobiasfactors} distinguish answer flipping from positional skew and show that aggregation can reduce flipping partly by stabilizing an incorrect answer. Their decomposition applies under balanced rotations, which our sampled permutations do not enforce. Likewise, the permutation metric of \citet{permutationlora} measures changes in candidate probabilities, whereas our analysis follows returned selections. We distinguish variation under repeated requests from variation under reordered menus, then test whether voting corrects errors and whether agreement identifies correct answers for Jev and the LLM baselines.

\section{Conclusion}
\label{sec:bias-conclusion}

A simple reordering test reveals a limitation of Jev-style System One models: stable accuracy and repeatable choices can conceal the loss of successful decisions. Across eight presentations of the primary collection, the two Jev versions' failure sets expand by 27.3--28.5\% while accuracy remains nearly unchanged. Reordering loses substantially more initial successes than fixed-order repetition, a gap also observed in highly repeatable Jev reproductions. The LLM comparison establishes no general Jev advantage, and the vulnerability persists at higher accuracy and with short menus. Voting offers limited correction for Jev, and unanimous errors remain undetected by agreement alone. The central requirement is to evaluate whether individual correct choices survive equivalent presentations. Establishing which design choices improve that survival requires controlled comparisons beyond the shared interface.

\bibliographystyle{colm2026_conference}
\begingroup
\def\UrlBreakPenalty{100}
\expandafter\def\expandafter\UrlBreaks\expandafter{\UrlBreaks\do\0\do\1\do\2\do\3\do\4\do\5\do\6\do\7\do\8\do\9}
\bibliography{references}
\endgroup

\clearpage
\appendix
\section*{A Survey of Jev-Style System One Models}

\section{Survey Overview}
\label{sec:introduction}

The main study shows that Jev's stable accuracy and high fixed-order agreement can coexist with substantial losses of correct choices under reordering. It leaves the responsible mechanisms unresolved. Jev-style methods share a candidate-distribution interface while differing in representation, learning, computation, and downstream use. This survey organizes those differences to identify what a controlled test of structural robustness would need to vary.

A candidate distribution can be extracted from a frozen language model or learned from labels, contrastive pairs, reasoning teachers, and policy rewards \citep{inner,llmasjev,jeval,openrlcd}. A direct response can also arise from intermediate exits, recurrent latent processing, or conditional reasoning \citep{fundm,sansi,startlux}. The output format therefore does not specify how a decision is learned or computed.

We organize the evidence around model construction, downstream workflows, and evaluation, building on prior distinctions among decision-model roles \citep{typedaudit,wild}. Appendices~\ref{sec:design} and~\ref{sec:execution} examine how distributions are learned and computed; Appendix~\ref{sec:workflows} examines what they control. Appendix~\ref{sec:evaluation} connects local prediction quality to task outcomes, where local savings need not improve overall success \citep{agents}. Appendix~\ref{sec:rethink} develops the resulting controlled comparisons.

\section{Decision Interfaces and Contribution Boundaries}
\label{sec:scope}

\subsection{The Decision Contract}

A decision interface changes the contract between a model and its caller. Given a state $s$, question $q$, and finite candidate set $\mathcal C$, the model reports
\begin{equation}
 p_\theta(c\mid s,q,\mathcal C),\qquad c\in\mathcal C.
 \label{eq:interface}
\end{equation}
Because the caller supplies candidates at inference, this contract extends beyond classification into a fixed training vocabulary. Choices, binary judgments, and ordinal scores can all expose such distributions. Constraining the response to the declared output space ensures its type; it does not ensure semantic correctness, calibration, or consistency across questions.

Official Jev combines this interface with a proprietary model and a service that reuses shared state across parallel questions. TypeSafe describes a new architecture and Reinforcement Learning for Calibrated Decisions (RLCD), but the public documentation available at the evidence cutoff does not disclose a reproducible architecture, training distribution, or exact objective.\footnote{Official launch: \url{https://typesafe.ai/blog/introducing-system-one-models-and-jev}; machine-learning primer: \url{https://docs.typesafe.ai/introduction/machine-learning-primer}.} Consequently, public alternatives reproduce aspects of behavior or explore independently specified designs. They do not establish that Jev's internal algorithm has been recovered.

Model version and service behavior are also distinct. Shared-state execution, candidate limits, scalar-score conventions, and calibration transformations can change behavior behind the same interface.\footnote{Official model specifications: \url{https://docs.typesafe.ai/models}.}

We cover publicly documented finite-candidate models, selected multimodal extensions, and downstream decision workflows available by October~19, 2026, identified through arXiv, citation tracing, official documentation, and linked releases. Appendix~\ref{sec:inventory} summarizes additional implementations.

\subsection{The Object of Contribution}
A paper's contribution depends on what its Jev call does (Table~\ref{tab:roles}). AutoTrust uses Jev to obtain training targets for a replacement model; Jev-Mem uses it to control memory operations. An integration study may instead evaluate an existing workflow. These roles can overlap: CLM and FunDM each provide a model recipe and a coding-run selector, JEVal provides a benchmark and SII, and LAVOIR extends Laya while defining an information-acquisition policy.

\begin{table}[!htb]
\centering
\small
\setlength{\tabcolsep}{4pt}
\renewcommand{\arraystretch}{1.18}
\begin{tabularx}{\linewidth}{@{}L{2.5cm}>{\raggedright\arraybackslash}X>{\raggedright\arraybackslash}X@{}}
\toprule
\textbf{Contribution} & \textbf{Question answered} & \textbf{Examples and boundaries}\\
\midrule
Model construction & How is a decision model obtained, and what does its inference compute? & Jeff, SII, LDM, AutoTrust; CLM/FunDM model recipes. Jev can be a target behavior, teacher, or baseline.\\
Downstream workflow & What does the model control in task execution or in learning another agent? & Jev-Mem, Ultrafast, LAVOIR interaction; run selectors and frozen-Jev RL assistance. The model need not be proprietary Jev.\\
Evaluation and analysis & Which capabilities, system outcomes, or usage patterns are supported by evidence? & JEVal benchmark, agent integration studies, coherence tests, ecosystem surveys. Evaluation is distinct from workflow invention.\\
\bottomrule
\end{tabularx}
\caption{Three levels of contribution. A work may contribute at more than one level.}
\label{tab:roles}
\end{table}

Existing surveys already recognize distinct uses of typed decision models. \citet{typedaudit} discuss roles including controllers, judges, and replacements for individual steps, while \citet{wild} distinguish application domains from decision purposes in public repositories. We build on that distinction and focus on how specific model mechanisms connect to specific workflow operations.

\section{Constructing Decision Distributions}
\label{sec:design}
\label{sec:routes}

Candidate probabilities depend on the readout, the information shared among candidates, the supervision target, and the gradient estimator. These choices determine how order can enter the prediction and which interventions can change it. We examine each before turning to inference in Appendix~\ref{sec:execution}.

\subsection{Decision Readout Before Specialized Training}
\label{sec:frozen}

\takeaway{obs:readout}{A direct decision interface can expose a frozen model's existing predictions.}
A language model's existing probabilities can be restricted and normalized over a candidate menu without updating its weights. This frozen readout provides a baseline for assessing what specialized adaptation adds, even when its predictions remain imperfect.

The simplest route reads an existing model's candidate-token logits. The frozen baseline in Inner Jev renders alternatives with answer letters, disables thinking, and applies a softmax over the corresponding original language-model head rows \citep{inner}. Writing these logits as $z_k$, its calibrated readout is
\begin{equation}
 p_k=\frac{\exp(z_k/T)}{\sum_{j=1}^{K}\exp(z_j/T)}.
 \label{eq:readout}
\end{equation}
No model weights need change, although fitting $T$ still requires calibration data. SemIf and several community implementations use closely related answer-logit extraction. This route is an essential control for attributing improvements to specialized training: comparison against an LLM that must generate probability strings measures a different inference procedure.

LLM-as-Jev removes the restriction to a fixed pool of single-token labels without altering the architecture or tokenizer \citep{llmasjev}. A prompt ending in \texttt{Best answer: [} assigns candidate $k$ the suffix \texttt{k]}. The closing bracket makes these suffixes prefix-free. For suffix tokens $c_{k,1:m_k}$, it computes
\begin{equation}
 s_k=\sum_{t=1}^{m_k}\log p_\theta(c_{k,t}\mid x,c_{k,<t}),
 \qquad p_k=\frac{e^{s_k}}{\sum_j e^{s_j}},
 \label{eq:suffix}
\end{equation}
where $x$ contains the shared context and all alternatives. Prefix caching and parallel teacher-forced scoring make this inexpensive, but the paper describes generally one to three short forward steps after prefill. ``No free-form answer generation'' is therefore more accurate than a universal single-pass claim. The same framework can also inherit explicit thinking from its backbone.

\begin{figure}[!htb]
\centering
\includegraphics{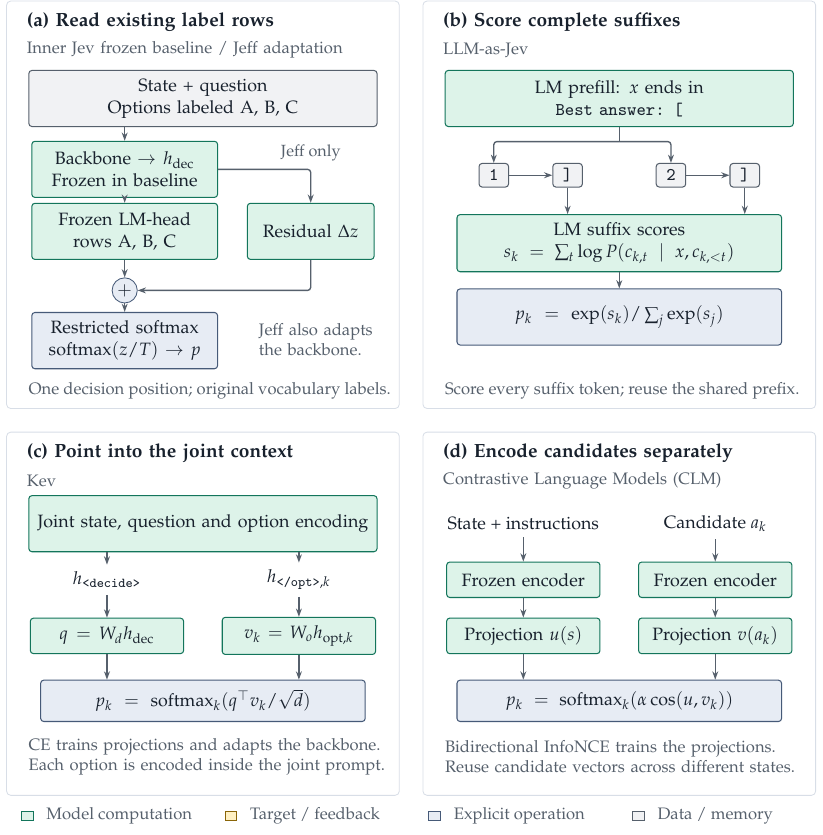}
\caption{\textbf{Four readouts for candidate probabilities.} (a) Vocabulary rows, (b) complete suffixes, (c) joint-context option states, and (d) separately encoded states and actions. Suffix token boundaries are illustrative. Across the mechanism figures, green denotes model computation, gold supervision or external feedback, slate operations, and gray data or memory. Labels specify training status. Sources and objectives appear in Appendix~\ref{sec:fitting}.}
\label{fig:readouts}
\end{figure}

\subsection{Candidate Representations, Interaction, and Reuse}
\label{sec:architecture}

\begin{figure}[!htb]
\centering
\includegraphics{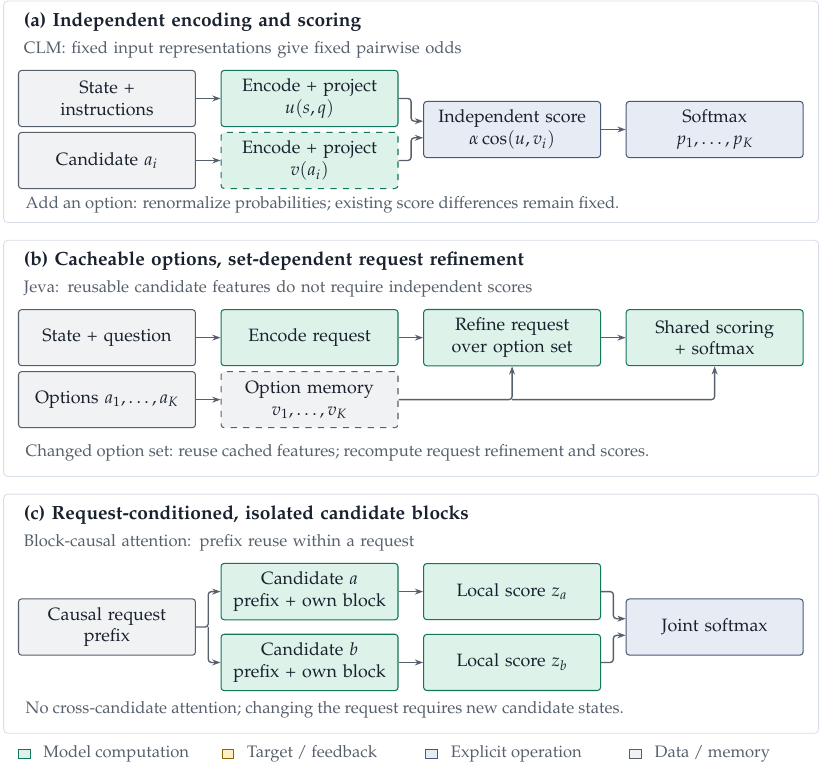}
\caption{\textbf{Candidate reuse and interaction are separate choices.} (a) Independent scoring, (b) set-dependent request refinement, and (c) request-conditioned candidate blocks. Arrows show dependencies; dashed outlines mark representations reusable with fixed inputs and weights. Odds statements fix request text, candidate text, and calibration.}
\label{fig:coupling}
\end{figure}

\takeaway{obs:interaction}{Reusable candidate representations can support set-dependent decisions.}
Candidate representations can depend on the request, rival options, and serialization position. Caching constrains their computation but not how later layers combine them: reusable options can support set-dependent decisions, while request-conditioned options can remain isolated from rivals (Figure~\ref{fig:coupling}). Order invariance and sensitivity to meaningful set changes are therefore separate properties.

\paragraph{Independent scores with reusable candidate representations.}
CLM independently encodes state--instruction and action inputs, then compares their projected representations. If these inputs and the temperature remain fixed, adding an alternative changes the softmax denominator but not the odds between two existing alternatives. MetaEncoder also uses a shared bi-encoder and contrastive training, but supports multimodal requests and updates the encoder \citep{metaencoder}. Its request-to-candidate InfoNCE is unidirectional, unlike CLM's bidirectional objective. Moreover, its closed-set request template can include the alternatives, which are randomly dropped during training. Candidate features remain request-independent, but the request representation need not be independent of the set. Thus a bi-encoder label alone does not establish invariant pairwise odds.

\paragraph{Reusable candidates with set-dependent decisions.}
Jeva independently encodes the request and each option, then refines only the request state against immutable option memory \citep{jeva}. Candidate representations exclude the current request, their position, and other options, enabling reuse across requests. A compact decision stack nevertheless allows the final request state to depend on the entire option set. Shared scoring and symmetric aggregation are designed to make the resulting probability vector permutation-equivariant, while allowing the relative ranking of existing options to change when the set changes. Jeva further separates competition among options from coverage: for closed-set probabilities $q_i$ and estimated coverage $\rho$, the open-set output is $p_i=\rho q_i$ with $p_{\varnothing}=1-\rho$. A diffuse $q$ expresses uncertainty among options; low $\rho$ expresses possible answer absence. Their empirical quality still depends on learning and evaluation.

\paragraph{Request-conditioned candidate blocks.}
Candidate-independent block-causal attention offers a third design \citep{blockcausal}. A causal request prefix is shared; each candidate block attends to that prefix and its own tokens, but not to other candidate blocks. Reset positions and a shared local score head aim to remove serialization-order effects. Candidate states therefore depend on the request and cannot generally be cached across different requests, even though their computation excludes other candidates. This is different from both independent bi-encoding and Jeva's immutable option memory. The reported residual permutation flips also distinguish a structural design objective from exact numerical behavior. Its frozen contrastive baseline uses a different training setup, so that comparison does not isolate attention structure alone.

Prism-1 uses the same separation of a shared causal prefix and isolated candidate blocks, but scores each candidate with the native language-model head's yes/no log odds at its final token \citep{prism}. A softmax across these scores produces the choice distribution; LoRA and candidate cross-entropy adapt the model without a separate learned scalar head. Every candidate starts at the same position offset. For a fixed prefix and deterministic computation, isolation and position reuse make scores independent of candidate serialization order and preserve pairwise odds when unrelated candidates are added. The latter is a restriction as well as an invariance: a task requiring an existing option's score to change with its rivals needs a different dependency structure.

\subsection{Supervision and the Target Distribution}
\label{sec:fitting}

\takeaway{obs:targets}{The supervision source determines what a fitted decision distribution represents.}
Labels describe outcomes in the training population, contrastive pairs distinguish matched inputs from chosen negatives, and teacher probabilities reflect the teacher's information and computation. Similar cross-entropy expressions can therefore fit different quantities.

\paragraph{Observed and mixed-label supervision.}
Jeff adds a residual decision head to the original letter readout and trains that head with backbone LoRA using cross-entropy \citep{inner}. The supervision mixes ordinary dataset labels, human uncertainty and ordinal annotations, and teacher-provided soft distributions. It is neither a pure hard-label recipe nor an RL method. Per-type temperature fitting follows training. The calibrated frozen model already provides a strong reference on the authors' decision benchmark, while their knowledge-oriented comparisons still favor Jev. These results motivate a strong baseline; they do not establish equivalence across workloads.

The adapted LLM-as-Jev variant instead preserves the architecture and candidate suffix representation \citep{llmasjev}. Its training loss uses a locally normalized candidate-prefix tree plus KL anchors that preserve the frozen model's behavior on legal/illegal token mass and outside decision positions. That locally normalized training distribution should not be confused with the globally normalized suffix likelihood in Equation~\eqref{eq:suffix}. The recipe uses public human-annotated training data rather than teacher-synthesized labels and studies both full adaptation and LoRA.

Kev illustrates supervised adaptation with a different readout. Its pointer head projects the decision position and each option boundary into a shared space and normalizes scaled dot products over options. The released recipe uses cross-entropy; smaller models use LoRA, while the documented 27B model uses full-parameter adaptation.\footnote{Official implementation and recipe: \url{https://github.com/jaredpalmer/kev}. Optional permutation and ordinal losses exist in the code but are not part of the documented released-model recipe.}

Intern-Decision instead retains the native vocabulary head and fits hard-label, full-vocabulary cross-entropy only at shifted decision positions \citep{interndecision}. Its input contains the complete ordered schema and an assistant JSON skeleton with \texttt{<decision>} placeholders. The logits immediately before each placeholder predict that field; earlier placeholders remain visible without feeding back gold or predicted answers. Inference restricts these logits to legal options and applies temperature scaling. The language parameters are adapted while the vision encoder and projector remain frozen. Training over the full vocabulary and restricting the inference readout are distinct operations, and the shared causal pass does not imply independence or order invariance across fields.

\paragraph{Modality-specific readouts.}
MJev-Spatial reuses pretrained query/key projections from the final full-attention layer to match a decision state to option-end states, rather than adding a prediction head \citep{mjevspatial}. Mean scaled inner products across query heads become logits for one candidate softmax. The retained language parameters are trained with cross-entropy while the vision encoder and merger are frozen. Because option states come from a causal sequence, later options can see earlier ones; shuffling options during training does not establish structural permutation equivariance.

Qwen-Audio-Jev uses a different design: a shared audio--language prefix state feeds three affine heads for 64 Choice indices, one binary Noul logit, and ten ordinal Score indices \citep{qwenaudiojev}. Candidate meanings are specified in the input, and each categorical head is normalized over the valid indices. Supervised training updates the audio encoder, adaptor, backbone, and heads using the observed primitive's loss: Choice cross-entropy, binary cross-entropy, or ordinal cross-entropy with cumulative-distribution and expected-index penalties.

\paragraph{Contrastive state--action alignment.}
Contrastive Language Models (CLM) use a frozen Qwen3-8B encoder with learned state and action projection heads.\footnote{Original technical blog: \url{https://contrastive-lm.notion.site}; official implementation and reference-model recipe: \url{https://github.com/Contrastive-LM/CLM}.} Matched state--action pairs supervise bidirectional InfoNCE: the state retrieves its matched action and the action retrieves its matched state among batch alternatives. At inference, state plus instructions and each candidate are encoded independently, and a softmax over scaled cosine similarities gives the candidate distribution. Independent candidate representations can be cached, although different question instructions change the state encoding. This is learned contrastive alignment rather than teacher-probability distillation.

\paragraph{Reasoning-endpoint distributions.}
SII-Decision trains a direct readout against a same-size backbone's reasoning-conditioned predictions \citep{jeval}. For each question, two reasoning chains yield two candidate distributions; their mean becomes the cached student target.\footnote{JEVal Section~6.2 specifies a nonthinking-readout fallback for invalid chains, while Appendix~F.4 describes excluding truncated, invalid, or top-option-disagreeing chains from added-source training data.} Cross-entropy to this target is combined with a KL constraint to the original nonthinking prediction, and ordinal tasks also use an expected-level penalty. The student updates language layers while retaining the original label head, and inference uses a single prefill with no generated rationale. This is offline reasoning-to-readout self-distillation, not student on-policy rollout distillation.

Where SII retains reasoning-endpoint probabilities, LDM constructs soft targets from sampled answers \citep{ldm}. In the primary experiments, its SPRD variant draws four reasoning answers from a frozen Qwen3.8-27B teacher for all student sizes and forms a vote histogram normalized by the number of parseable answers. SPRD already requires the top vote share over all four samples to reach $0.75$; SPRD-C additionally requires the majority to agree with the label intended during data construction. This checks consistency, not independent correctness. The LDM objective uses supervised NLL on one record type and weighted teacher-to-student KL on another, rather than applying both losses to every example.

Endpoint averaging and answer counting differ as estimators, but need not estimate different quantities. If both use the same rationale distribution $\rho(r\mid x)$ and answers are sampled categorically from the same endpoint distribution $q(\cdot\mid x,r)$, then
\begin{equation}
 \mathbb E_{r\sim\rho}[q(y\mid x,r)]
 =\mathbb E_{r\sim\rho,\,a\sim q(\cdot\mid x,r)}[\mathbf1\{a=y\}].
 \label{eq:teacher-marginal}
\end{equation}
Averaging conditional probabilities removes the final-answer sampling noise under these assumptions. Different teachers, prompts, decoding rules, or sample filters can change the target marginal; Equation~\eqref{eq:teacher-marginal} does not establish equivalence between the actual SII and LDM datasets.

\paragraph{Full-depth predictions as a teacher.}
Teacher computation can also differ in depth rather than in the use of reasoning traces. FunDM freezes the backbone and trains a decision head on an intermediate hidden layer to match the full-depth model's distribution using $\mathrm{KL}(p_{\mathrm{teacher}}\|p_{\mathrm{student}})$ \citep{fundm}. Inference can stop before the remaining layers, although the new head still requires training.

\paragraph{Black-box imitation of the deployed service.}
AutoTrust-JEV-27B uses output distributions obtained from TypeSafe Jev~1.13 as its main supervision stream.\footnote{Author model card and recipe: \url{https://huggingface.co/autotrust/JEV-27B}.} Its training corpus also contains programmatic labels and downweighted uniform-placeholder targets. The documented recipe fits LoRA adapters and a decision head using target-to-student KL, with an ordinal ranked probability score (RPS) term. The service-target component is behavioral imitation of a black-box model. Agreement with that teacher is a different outcome from correctness against independent labels, and it does not identify the teacher's internal training algorithm. External-teacher models such as Eikos and Torchcast provide further distribution-fitting variants, catalogued in Appendix~\ref{sec:inventory}.

\begin{figure}[!htb]
\centering
\includegraphics{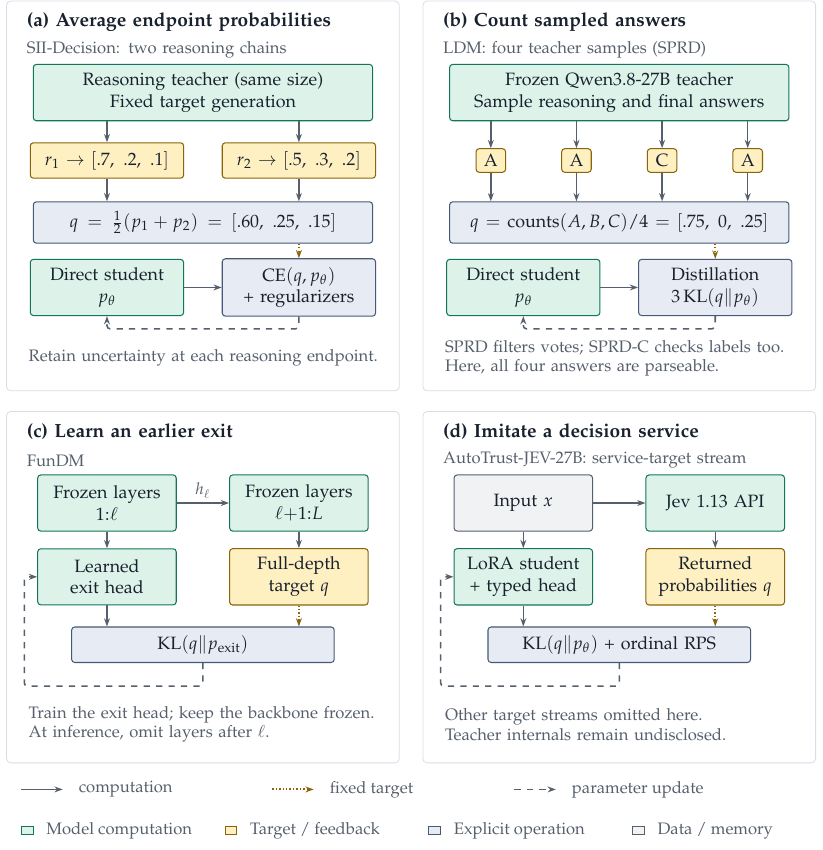}
\caption{\textbf{Four sources of teacher targets.} (a) Averaged endpoint probabilities, (b) sampled-answer votes, (c) full-depth predictions, and (d) AutoTrust service outputs, excluding its other target sources. Vectors in (a,b) are illustrative; (b) assumes four parseable samples. Dotted gold arrows carry fixed targets; dashed arrows mark updated components. Student inference omits teacher computation.}
\label{fig:targets}
\end{figure}

These teacher targets differ in the information and computation they retain (Figure~\ref{fig:targets}).

\paragraph{Workflow-specific supervision.}
Some models are trained to support a particular workflow. LAVOIR adds slot representations and a value-of-information head to Laya, trained with decision cross-entropy and value regression \citep{lavoir}. The final recipe sets the inherited RL weight to zero. The model learns both a decision distribution and estimates used by the ask/decide/handoff policy in Appendix~\ref{sec:clarification}.

AgentDecision-4B, introduced in JevAgentHarness, trains the decision component on agent states \citep{harness}. Its Qwen3-4B LoRA recipe includes post-error states from mistake replay. When Jev's top option matches the gold label, the supervised target mixes $0.3$ gold one-hot mass with $0.7$ teacher probabilities; otherwise it uses the gold label alone. Cross-entropy plus a Brier term precedes closed-loop GRPO with a fixed LLM, task-success and cost/error rewards, and a KL anchor to the supervised model. Here environment feedback updates the decision model itself. Appendix~\ref{sec:downstream-learning} examines the different role of a fixed Jev service guiding another learner.

\FloatBarrier
\subsection{Statistical Objectives and Gradient Credit}
\label{sec:policy}

\takeaway{obs:credit}{The probability objective and the random variable receiving credit are separate choices.}
A proper probability score defines an objective for truthful reporting; a gradient estimator determines how its training signal reaches the parameters. For a distribution $p$ and outcome $y$, log score $\log p_y$ and negative Brier score $-\sum_k(p_k-\mathbf1[k=y])^2$ reward truthful probability reports in expectation. A differentiable readout can optimize either score directly. By contrast, rewarding only a sampled answer's correctness maximizes $\sum_k p_k q_k$ under outcome distribution $q$, generally favoring a mode of $q$ rather than reporting $q$. Thus the word ``RL'' identifies neither the statistical property being optimized nor an advantage over direct proper-score fitting.

\begin{figure}[!htb]
\centering
\includegraphics{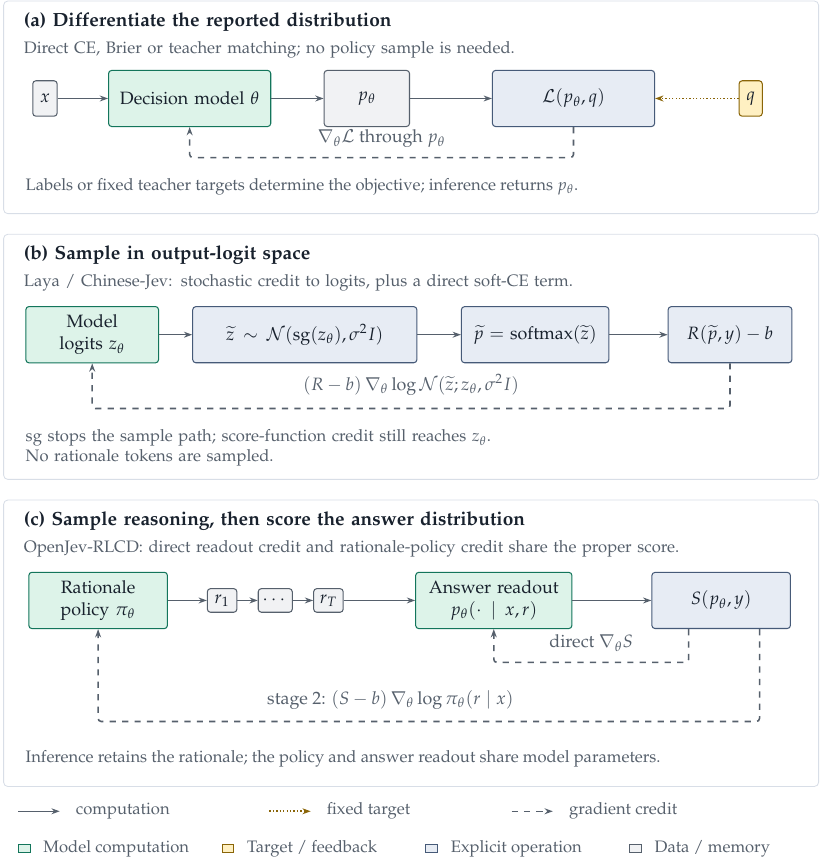}
\caption{\textbf{Three paths for learning signals.} (a) Direct differentiation, (b) output-space score-function credit supplementing soft cross-entropy, and (c) direct answer-readout gradients with second-stage rationale-policy credit. Here $b$ is a baseline, $q$ a fixed target, and $\operatorname{sg}$ stop-gradient. Schematics omit centering, normalization, and auxiliary regularizers; they compare mechanisms, not measured estimator advantages.}
\label{fig:gradients}
\end{figure}

Figure~\ref{fig:gradients} contrasts direct differentiation, output-space perturbations, and rationale-policy credit. OpenJev-RLCD, the rationale-policy method depicted here, retains generated reasoning at inference; this is a property of that method rather than a requirement of every rationale-based training procedure.

\paragraph{Output-space policies.}
Laya and Chinese-Jev perturb output logits and apply policy-gradient updates in that space; Chinese-Jev combines soft cross-entropy with Gaussian perturbations around detached logits and a group-baseline estimator \citep{chinese}. This does not require a generated chain of thought. The decider-derived this-that-model combines CE and Brier terms with a sampled-outcome Brier reward and a control-variate estimator \citep{thisthat}. Another output-policy variant, decider-2B v10, uses PPO over environment actions with terminal returns and an auxiliary proper-log-score belief objective; the documented 4B and 35B releases do not use that RL stage.\footnote{Version-specific recipe: \url{https://github.com/Mapika/decider/blob/main/docs/RL.md}.} These methods apply stochastic policy updates in output space rather than over generated reasoning. Their estimators, stop-gradient choices, and rewards determine what those updates contribute beyond direct fitting.

Qwen-Audio-Jev makes this distinction explicit with an importance-weighted, pairwise sampled-action estimator for negative Brier score and ordinal RPS \citep{qwenaudiojev}. Its exploration mixture, leave-one-out baseline, and detached weights yield the direct proper-score gradient in expectation under the stated construction. Thus the estimator need not define a different probability target from direct fitting. This equivalence does not establish an empirical advantage in variance, efficiency, or calibration.

\paragraph{Reasoning-token policies.}
OpenJev-RLCD first calibrates an answer distribution conditioned on sampled rationales, then adds REINFORCE credit to the rationale policy using the same proper score \citep{openrlcd}. If
\begin{equation}
 J(\theta)=\mathbb E_{r\sim\pi_\theta(\cdot\mid x)}
       [S(p_\theta(\cdot\mid x,r),y)],
\end{equation}
then its two relevant gradient paths are the differentiable readout term and the score-function term through $\pi_\theta(r\mid x)$. Stage one uses the former; stage two adds the latter, with a baseline and KL regularization. Its main experiments score each rationale's distribution separately and retain explicit rationales at inference. It is an independent RLCD interpretation, not a disclosed implementation of proprietary Jev.

StartLux combines instant decision learning with an AutoThink variant that allocates reasoning conditionally \citep{startlux}. Its instant model makes full-attention layers bidirectional while retaining causal linear-attention layers. RLCD$^2$ uses a frozen reference model's before/after probability signal to train a reasoning-token policy. Within groups containing only nonempty traces, the initial-score term cancels under reward centering; empty traces receive zero reward, so this cancellation does not extend to arbitrary mixed groups. PostHog's Jeeves is another reasoning-policy implementation: its repository combines supervised learning, CISPO-style updates, a pointer head, and a diffusion drafter. Its documented reward uses correct-answer probability with a length penalty, which should not automatically be called a proper scoring rule.

\FloatBarrier
\section{Executing Decision Models}
\label{sec:execution}

\begin{figure}[!htb]
\centering
\includegraphics{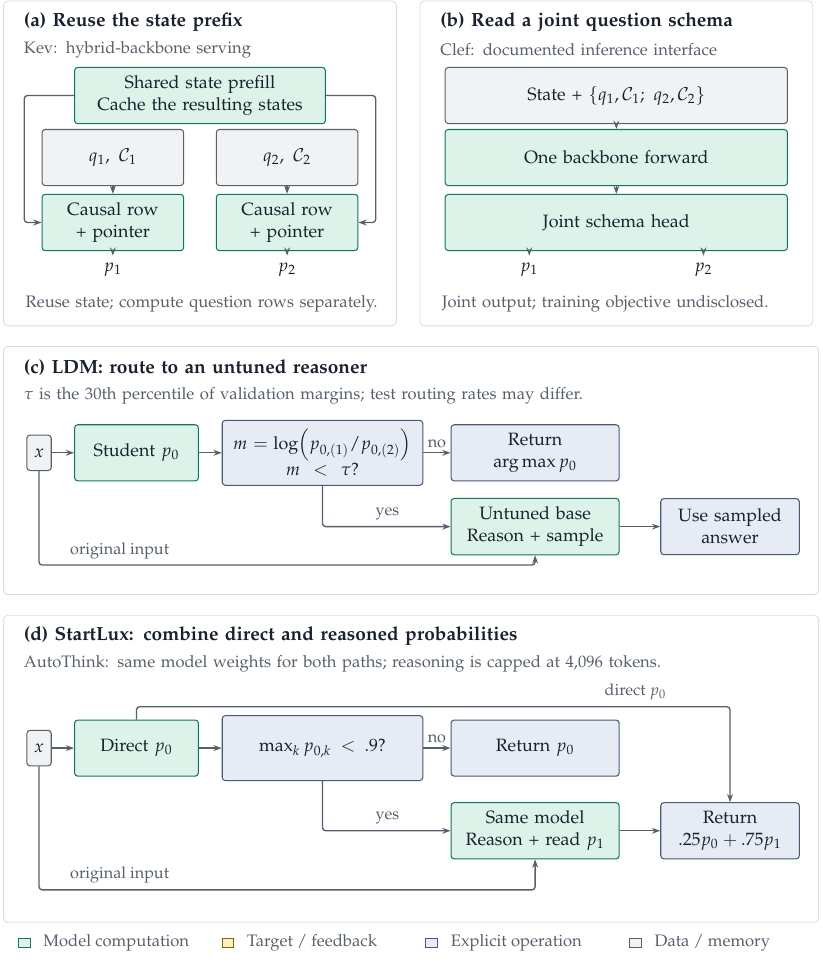}
\caption{\textbf{Direct outputs permit different execution paths.} (a) Kev cache reuse, (b) Clef joint schema readout, (c) LDM routing to untuned-base-model reasoning, and (d) StartLux fusion with the same weights on both paths. $p_{0,(1)}$ and $p_{0,(2)}$ are the two largest readout probabilities. Arrows show inference flow, without parameter updates.}
\label{fig:execution}
\end{figure}

The readouts and learning recipes above leave several execution choices open. A direct choice can follow different computational paths and undergo serving transformations. Figure~\ref{fig:execution} contrasts computation shared across questions (a--b) with computation adapted within a decision (c--d). Both affect cost, while routing and output transformations can also change the returned prediction.

\subsection{Direct Outputs and Adaptive Computation}
\label{sec:compute}
\takeaway{obs:computation}{The absence of a returned rationale does not imply a fixed computation budget.}
A direct student may omit its teacher's reasoning, an intermediate head may omit later layers, and a recurrent model may perform additional hidden steps without generating text. Conditional reasoning introduces another possibility, mixing direct and deliberative requests behind one interface. These choices change different resources, so generated-token count alone cannot compare their computational demands.

\paragraph{Conditional reasoning changes both the path and the prediction.}
LDM routes low-margin items to the student's untuned base for one reasoning rollout and adopts its answer; the direct branch keeps the readout's answer (Figure~\ref{fig:execution}c; \citealp{ldm}). Its threshold is the validation margins' 30th percentile, which does not fix the routed fraction on unseen data. StartLux AutoThink instead uses the same model weights for the initial causal readout, reasoning, and endpoint readout (Figure~\ref{fig:execution}d; \citealp{startlux}). With a maximum initial probability below $0.9$, it generates up to 4,096 reasoning tokens and returns $0.25p_0+0.75p_1$; otherwise it returns $p_0$. These routes differ in the model called, the trigger, and the returned prediction.

\paragraph{Hidden computation and selective reasoning.}

Latent loops allocate extra computation without generating a textual rationale. SanSi repeatedly applies a recurrent backbone and reads a decision distribution after each loop \citep{sansi}. The base weights are frozen, while shared LoRA adapters and loop-specific readout corrections are learned. Cross-entropy and Brier losses supervise the loop outputs, with gradients passing through preceding loops. The paper trains with eight loops and evaluates different loop counts. Latent computation thus varies separately from parameter count and generated-token count, and additional loops need not improve every task.

Selective Reasoning instead exposes three control decisions: when to start generation, when to read a partial rationale, and when to accept the current answer \citep{selective}. A reader is trained on empty and partial reasoning prefixes using a frozen backbone. The start gate estimates correction benefit relative to regression risk, the read scheduler limits costly reader calls, and the accept rule uses the reader's distribution. An accepted prefix answer cancels generation only if generation is still active. If no prefix answer is accepted, a terminal read returns an answer without a certainty requirement. Thus the entropy threshold governs early acceptance rather than guaranteeing the confidence of every returned answer. The threshold itself measures distribution concentration, not an independently measured probability of correctness.

FastJEV changes the execution of an existing decision model without weight training \citep{fastjev}. It reuses shared context and recurrent states, forks candidate-prefix caches, and selects a candidate-path layer-skipping mask using relative-score perturbations on unlabeled calibration inputs. These mechanisms preserve the context and candidate inventory while changing the work performed on candidate paths. Prefix reuse can reduce computation yet increase latency through execution overhead. Its results therefore support a serving analysis under specified workloads, rather than a universal latency improvement. The OmniJev model evaluated in that paper should not be conflated with the separately named OneJev repository.

Comparing these paths requires accounting for backbone depth, reader calls, generation, caching, and scheduling. Returned-answer length alone does not measure their cost.

\paragraph{A single-pass primitive can form a sequential predictor.}
MJev-Spatial illustrates a further distinction between the primitive and the completed prediction \citep{mjevspatial}. A categorical answer uses one decision, while spatial point prediction uses three successive nine-way cell choices. The first two choices determine new crops from the original image, which are re-encoded for the next stage. Training packs teacher-forced stages into one backbone pass, but inference must follow the predicted crops sequentially. The spatial quantization bound assumes the correct cell choices; it is not a bound on model prediction error. Single-pass training, single-pass primitive scoring, and single-pass task inference are therefore separate claims.

\subsection{Serving and the Returned Prediction}
\label{sec:release}
\takeaway{obs:serving}{Serving choices can change prediction semantics as well as execution cost.}
The model head is only one stage of a deployed prediction. Postprocessing can move its probability mass, scalar reduction can retain different information, and question batching can change how computation is shared. Behavior observed through a service can therefore reflect transformations applied after training as well as the learned distribution.

\paragraph{Sharing computation across questions.}
The same typed interface can expose different question-processing graphs. Kev's documented hybrid-backbone implementation processes separate causal question rows and can reuse a state cache in serving (Figure~\ref{fig:execution}a). Cloudflare Clef instead uses a joint schema head and describes one forward pass for all questions (Figure~\ref{fig:execution}b). Jeff's released server evaluates questions serially, while official Jev describes shared-state parallel question processing. These differences determine how computation is shared even when each question returns the same output type.

Documented readouts do not always identify a training recipe. Perplexity's decider v1.1 describes a separate 255-row readout and saved temperature calibration, but the public model cards for decider and Clef do not specify their training objectives (Appendix~\ref{sec:inventory}). JevK5 further describes a one-pass readout only up to its candidate limit and a multi-pass tournament beyond it. Candidate limits and question organization are therefore necessary conditions for interpreting an execution claim.

\paragraph{Transforming the returned prediction.}
Jeff illustrates how postprocessing can change the apparent decisiveness of a model. Its default binary ``commitment'' transformation moves intermediate probabilities beyond the benchmark's abstention interval. The returned distribution therefore differs from the temperature-calibrated distribution before this transformation, and an evaluation of calibration must specify which one it measures. Reduced abstention under this rule cannot be attributed to the cross-entropy training objective alone.

SII returns the expected ordinal level, whereas Jeff's released scalar field returns an argmax level. Identical categorical distributions can therefore yield different scalar answers.

Table~\ref{tab:routes} relates learning recipes to their readouts and public artifacts.

\begingroup
\small
\setlength{\tabcolsep}{3pt}
\renewcommand{\arraystretch}{1.22}
\begin{longtable}{@{}L{2.15cm}L{4.05cm}L{3.95cm}L{3.1cm}@{}}
\caption{Model construction, readout, and public artifacts. Rows distinguish training stages and model variants.}\label{tab:routes}\\
\toprule
\textbf{Model / recipe} & \textbf{Target and update} & \textbf{Readout and inference} & \textbf{Public evidence}\\
\midrule
\endfirsthead
\multicolumn{4}{l}{\small Table~\thetable\ continued}\\
\toprule
\textbf{Model / recipe} & \textbf{Target and update} & \textbf{Readout and inference} & \textbf{Public evidence}\\
\midrule
\endhead
\bottomrule
\endfoot
Inner Jev / Jeff \citep{inner} & Frozen baseline; adapted model uses mixed-label CE, LoRA and residual head; temperature fitting & Up to 16 letter labels; one forward per question; released server runs questions serially & Paper, weights, serving code and OpenDecisions\\
LLM-as-Jev \citep{llmasjev} & Frozen suffix scoring; optional candidate-tree NLL with KL anchors on human labels & Original architecture; short multi-token suffix scoring; optional inherited thinking & Paper describing scoring and optional adaptation\\
CLM & Bidirectional InfoNCE on matched state--action pairs; frozen backbone, learned projections & Independently encoded state and candidates; softmax of scaled cosine scores; reusable embeddings & Technical blog, heads, inference and fine-tuning code\\
SII-Decision \citep{jeval} & Cached reasoning-endpoint distributions; CE, reference KL and ordinal loss & 255 native single-token labels; one prefill; no rationale; $T=1$ & Paper, training/inference code, four models and target datasets\\
LDM \citep{ldm} & Hard-label variants; SPRD matches concentrated teacher votes; SPRD-C also checks label agreement & Direct readout; optional untuned-base reasoning below validation margin's 30th percentile & Paper; repository states ``Code coming soon'' at the survey cutoff\\
FunDM \citep{fundm} & Teacher-to-student KL; train intermediate head, freeze backbone & Early-layer decision head; remaining backbone layers can be skipped & Paper and implementation\\
AutoTrust-JEV-27B & Jev~1.13, programmatic and downweighted placeholder targets; KL + ordinal RPS; LoRA and head & Direct candidate readout; larger menus extrapolate beyond trained range & Weights, inference code and detailed model card\\
Chinese-Jev \citep{chinese} & Soft CE plus logit-perturbation policy gradient, following Laya & Direct decision readout; output-space sampling during training & Paper and public model/repository sources\\
this-that-model \citep{thisthat} & CE + Brier; sampled-outcome Brier estimator with control variate & Direct finite-candidate readout; follows decider & Paper and linked implementation\\
OpenJev-RLCD \citep{openrlcd} & Proper-score readout fitting, then rationale REINFORCE + KL & Generates a rationale, then reads answer-token probabilities; fitted temperature & Paper, training code and result files\\
StartLux \citep{startlux} & Instant learning plus RLCD$^2$ reasoning-policy optimization & Instant: bidirectional full-attention layers; AutoThink: conditional reasoning & Paper and public release links\\
Jeeves & SFT + CISPO-style reasoning updates; correct-answer probability with length penalty & Pointer head and diffusion drafter; reasoning remains part of the design & Public training implementation\\
\end{longtable}
\endgroup

\FloatBarrier
\section{Decision Models in Downstream Workflows}
\label{sec:workflows}

Execution determines the cost and form of a returned decision; its consequence depends on where it intervenes in a task. We distinguish control of an ongoing trajectory, selection among completed outcomes, acquisition of evidence, and assistance to another learner. These roles carry different error and recovery costs, even when they use the same decision model.

\subsection{Controlling a Trajectory or Selecting Its Outcome}
\takeaway{obs:intervention}{Action control and completed-run selection affect different parts of a trajectory.}
Online selection changes which state the agent reaches next. Completed-run selection instead chooses among trajectories whose generation cost has already been incurred. Both can use a fast scorer, but only the former changes the actions within an individual run. This distinction determines which generator, candidate budget, and executor must remain fixed to attribute a downstream gain to the decision component.

\paragraph{Online action selection.}

\begin{figure}[!t]
\centering
\includegraphics{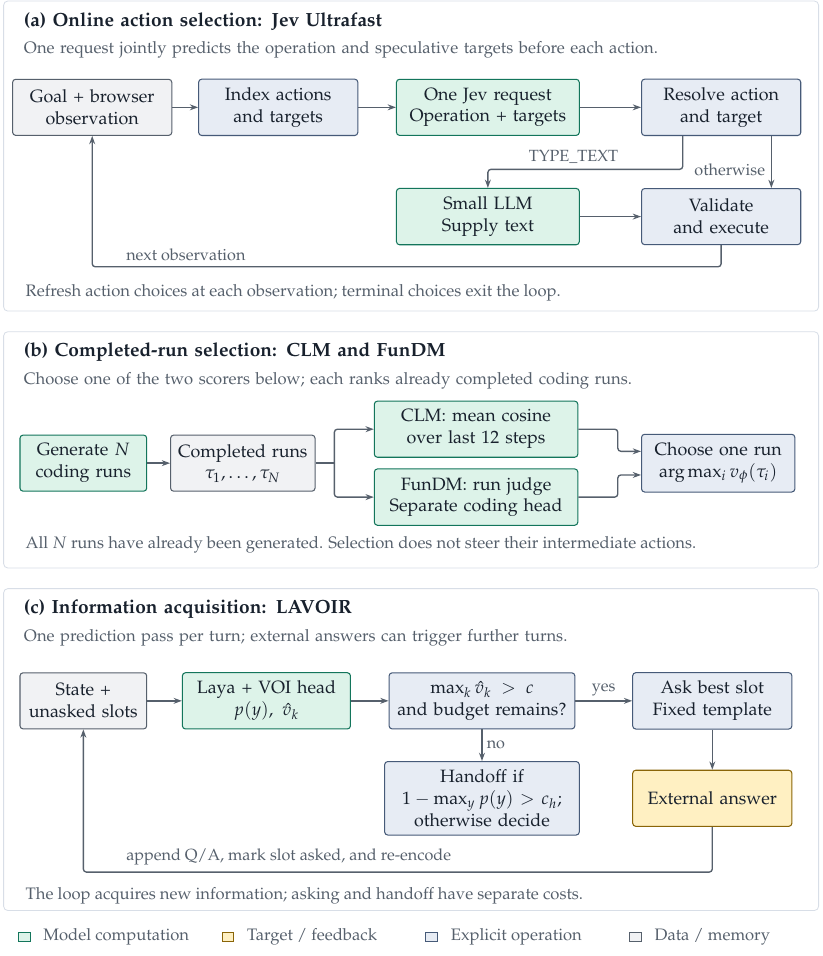}
\caption{\textbf{Decisions intervene at different workflow stages.} (a) Online action selection, (b) completed-run selection, and (c) repeated clarification, handoff, or decision. Arrows show runtime flow. Panel (b) scores completed runs rather than controlling their individual actions; CLM includes mean score aggregation. Colors follow Figure~\ref{fig:readouts}; sources appear in the corresponding subsections.}
\label{fig:workflows}
\end{figure}

Jev Ultrafast constructs a dynamic action space from the observed browser state.\footnote{Primary implementation: \url{https://github.com/browser-use/jev-ultrafast}, especially ``The action space'' and ``Evidence and limits.''} Each observation supplies an indexed table of elements. A single Jev request asks for both an operation and speculative targets for the supported operation types. The executor uses only the target corresponding to the selected operation (Figure~\ref{fig:workflows}a). A small LLM supplies text when the operation requires typing. Ordinary code checks target compatibility and page freshness, executes the action, and obtains the next observation.

The division of labor depends on enumerating actionable choices from the environment. The decision model selects within that interface, while generation remains available for open-ended content. Selection of a \texttt{DONE} action still needs an independent outcome check. The repository reports a small set of demonstrations and repeated runs on one browser task; these establish an implemented loop, not general reliability or an isolated effect of the decision model.

JevCoder exposes a related boundary in code repair \citep{jevcoder}. Its components recommend code snippets, guide action types, and judge whether to submit or continue. When Jev selects an action type, an LLM still generates the specific command. A judge that blocks a command does not thereby supply a better one, and some blocking interventions hurt performance. The full-workflow experiments are separate from component studies on selected success/failure panels; they should not be pooled as a single task-success estimate. Each full-workflow task receives one run per condition, which limits what can be inferred about repeated-run variability.

\paragraph{Decision gates within a larger controller.}
A decision gate can replace a local choice, advise a generator, reject an action, trigger extra work, or escalate uncertain cases. JevAgentHarness implements these roles at tool selection, bounded arguments, memory operations, reflection, stopping, and model routing \citep{harness}. Its deployment cascade compares the maximum option probability with thresholds fitted for each decision point using held-out shadow trajectories and intervention costs. Free-form arguments still require an LLM. The gate's confidence and the harness's policy jointly determine behavior; a threshold alone does not establish safe execution. The same work also trains a decision model and evaluates controllers, which are distinct contributions (Appendices~\ref{sec:fitting} and~\ref{sec:evaluation}).

Jev-LiteMAS specializes this division of labor to a predefined pool of generative experts \citep{litemas}. Jev selects the next expert or, when eligible, STOP, and screens artifacts for acceptance or escalation. On escalation, an LLM reviewer supplies a verdict and textual feedback; an expert revises the artifact, which is then screened again. The decision model does not generate repair instructions, and local artifact acceptance is distinct from stopping the entire workflow. Savings therefore depend on the escalation frequency and on whether the resulting artifacts still satisfy the task.

Jarvis applies typed control to an asynchronous task ledger \citep{jarvis}. Runtime code constructs eligible operations tied to task identifiers and versions; the decision model selects among them from the transcript, history, and ledger. Low-confidence cases delegate to a worker without speculative ledger mutation. Code enforces one live version, rejects stale results, revalidates plans, and requires a playback receipt before marking speech as delivered. These invariants belong to the executor, not to the model's probability estimates. The paper also trains a separate audio LISTEN/DIRECT/DELEGATE gate; that learned gate and the downstream ledger policy should not be treated as one training mechanism.

A penetration-testing harness study illustrates how several decision roles can coexist in one domain \citep{pentest}. Typed decisions support finding adjudication, severity assessment, agent pruning, and bounded confirmation, while the harness retains evidence gates and execution. Its exploratory comparison uses one run per condition on one target, and subsequent retests change the harness. The study demonstrates this division of labor, but the separate effects of the decision model and harness remain unresolved. Establishing a benefit would require comparing the same task decomposition and execution policy with alternative decision components.

\paragraph{Selecting completed runs.}

CLM and FunDM move the intervention from actions within a run to selection among completed coding-agent runs. If the generator produces trajectories $\tau_1,\ldots,\tau_N$, a scorer chooses
\begin{equation}
 \hat\imath=\arg\max_{i\in\{1,\ldots,N\}}v_\phi(\tau_i).
 \label{eq:runselector}
\end{equation}
The generation cost of all $N$ candidates remains part of the workflow. A cheap verifier may reduce selection overhead, but this alone does not make the total procedure cheaper than generating one run.

CLM's released coding verifier averages state--action cosine scores over the last twelve trajectory steps by default.\footnote{Official selector implementation: \url{https://github.com/Contrastive-LM/CLM/blob/main/evaluation/bon_eval.py}. The coding evaluation uses domain adaptation; it is distinct from the general typed-question interface.} This aggregation is a use of the contrastive scorer, not the softmax over a caller's candidate menu described in Appendix~\ref{sec:fitting}. FunDM uses a separately trained coding-agent head to judge a whole run \citep{fundm}. Its MCQ head and coding head are distinct, and the repository distinguishes the whole-run judge from an action-trained FunDM-CE variant.\footnote{FunDM implementation and reproduction notes: \url{https://github.com/Ujjawal-K-Panchal/FunDM}.} Figure~\ref{fig:workflows}b therefore shows a shared selection operation with different scoring mechanisms. It does not imply shared heads, training data, or directly comparable reported performance.

\subsection{Changing the Evidence Available to a Decision}
\takeaway{obs:information}{Clarification and memory control alter the information available to subsequent decisions.}
The preceding workflows use the available evidence to choose behavior or an outcome. Clarification obtains an external answer, while memory control changes how past evidence is organized and retrieved. These operations can improve a later decision without changing the decision model's parameters. Their benefits and costs therefore depend on the information acquired, as well as the quality of the local probability report.

\subsubsection{Acquiring Missing Information}
\label{sec:clarification}

LAVOIR predicts the value of acquiring missing information. It extends the open Laya model so that a forward pass returns both a temperature-scaled decision distribution $p$ and values $\hat v_k$ for unasked information slots \citep{lavoir}. For a gold decision $y^*$, the target and bounded prediction are
\begin{equation}
 t_k=\hat p_k(y^*)-\hat p_0(y^*),
 \qquad \hat v_k=\operatorname{stopgrad}\!\left(1-\sum_y p_y^2\right)\sigma(h_k),
 \label{eq:clarification-value}
\end{equation}
where $\hat p_0$ and $\hat p_k$ come from the same frozen, temperature-scaled snapshot before and after a sampled answer to slot $k$. The value head produces $h_k$, and $\sigma$ is the sigmoid function. The detached Gini factor bounds predicted values when the current decision distribution is concentrated. Training regresses $\hat v_k$ toward $t_k$, with both normalized by the target standard deviation, alongside decision cross-entropy. The final recipe sets the inherited policy-gradient weight to zero. Multiple complete profiles for a partial message make the regression estimate an expected gain under the sampled answers. Thus the learned value is a proxy whose validity depends on the frozen predictor and answer distribution. This construction is a model contribution, while the use of its outputs defines the interaction.

At runtime, the policy asks about $k^*=\arg\max_k\hat v_k$ when $\hat v_{k^*}>c$ and the question budget remains. The question comes from a slot template. Its answer is appended to the state, which is then re-encoded; an already asked slot is not repeated. If no question is taken, the policy hands off when $1-\max_y p_y>c_h$ and otherwise returns the most probable decision (Figure~\ref{fig:workflows}c). Here $c$ and $c_h$ are the asking and handoff thresholds. Single-pass prediction within a turn does not make the entire interaction a single-pass procedure.

Clarification changes the source of information available to the model. LDM and StartLux conditional thinking process the existing input through an internal rationale, whereas LAVOIR obtains an external answer. It also goes beyond deferring low-confidence cases by ranking particular missing fields by predicted benefit. The evidence is strongest for synthetic multi-turn dialogues, with limited real single-exchange tests and one training seed. Comparisons against slower value-of-information computation and an LLM agent remain incomplete, leaving open whether the learned acquisition policy offers a better quality--cost trade-off across interaction settings.

\FloatBarrier
\subsubsection{Controlling Memory Construction and Retrieval}
\label{sec:memory}

\begin{figure}[!htb]
\centering
\includegraphics{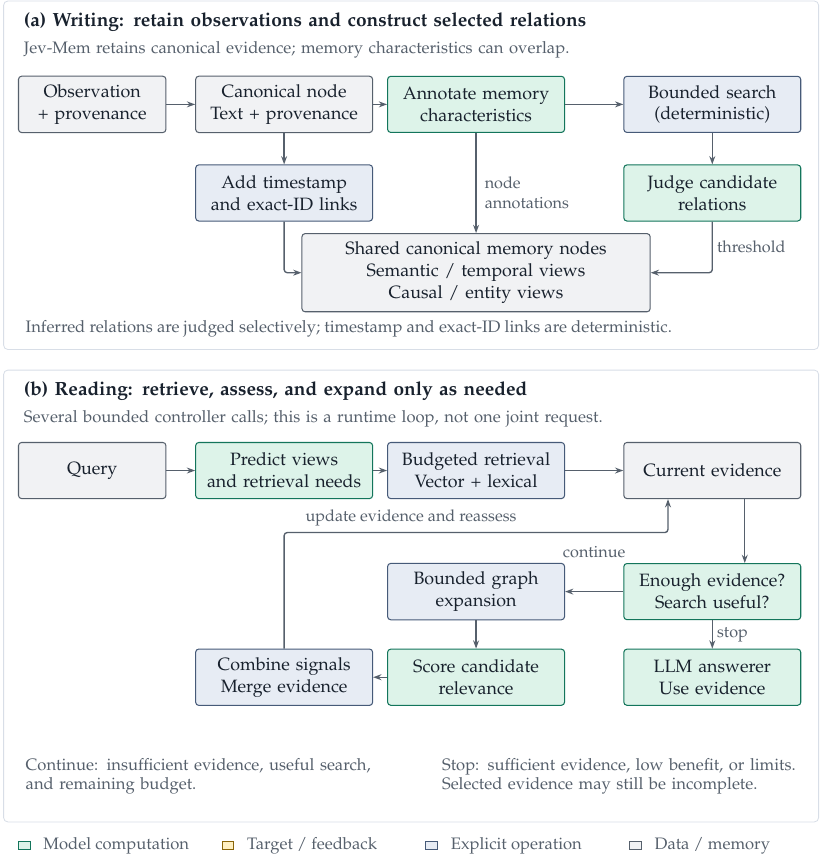}
\caption{\textbf{Jev-Mem controls memory writing and retrieval} \citep{jevmem}. (a) Writing preserves observations and judges relations. (b) Retrieval iteratively acquires evidence before answer synthesis. Four graph views share memory nodes. Arrows span multiple operations and calls; colors follow Figure~\ref{fig:readouts}.}
\label{fig:memory}
\end{figure}

Jev-Mem places a decision controller between stored observations and an LLM answerer \citep{jevmem}. Its contribution is the organization of memory operations around fast typed decisions, rather than a new recipe for training the controller. Shared canonical memory nodes participate in semantic, temporal, causal, and entity relation views. Figure~\ref{fig:memory} separates writing this memory from retrieving evidence, because the decision objects and feedback loops differ.

\paragraph{Writing preserves observations and judges relations.}
Incoming observations retain their original text and provenance. The controller annotates overlapping memory characteristics, while deterministic vector, lexical, entity, and time search produces a bounded set of candidate neighbors. The controller then judges inferred relations; thresholds determine which edges are added. Reliable timestamp ordering and exact identifiers can be handled directly by code. Thus the normal write path does not ask the model to irreversibly discard observations before their future relevance is known. Maintenance may mark redundancy, contradiction, or obsolescence while retaining original evidence, and textual abstraction can be escalated to an LLM.

\paragraph{Retrieval adapts both where to search and when to stop.}
Given a query, the controller predicts relevant relation views, multi-hop needs, and recency, which guide budget allocation. Hybrid vector and lexical retrieval supplies anchors. During retrieval, the controller assesses whether accumulated evidence is sufficient and whether further expansion is likely to help. If needed and within budget, bounded graph expansion proposes candidates, which are scored and added to the evidence set. The loop stops on sufficient evidence, low expected additional benefit, or a resource limit. Selected memories then go to the LLM for answer synthesis. The controller affects evidence acquisition; it does not replace the final generative answerer.

The reported evaluation uses LoCoMo with GPT-4o-mini as the answer model and LLM-as-a-Judge scoring. These results concern Jev-based control on LoCoMo and do not evaluate the Laya controller or LongMemEval configuration provided by the implementation.\footnote{Jev-Mem implementation: \url{https://github.com/libingzheren/Jev-Mem}.} Separating write and read control is particularly useful for attribution: changing stored relations can affect later answers even if the query-time controller is held fixed.

\paragraph{Evidence expansion without a preconstructed graph.}
SAG2 instead starts from a local SQLite FTS5/BM25 index, without corpus-time embeddings or a neighbor graph \citep{sag2}. Jev makes a binary Noul judgment about each passage's necessary contribution to a supported answer chain, conditioned on the retrieved evidence. This is not a single global Choice over passages. Ordinary code extracts lexical expansion terms from scored seed passages; retrieval and evidence-conditioned rescoring can repeat within a bounded budget. Local diversification then produces the final evidence set. The experiments evaluate retrieval rather than generated answers, and development includes MuSiQue and a synthetic enterprise corpus. The local index does not make the complete system local, because the evidence scorer uses the hosted Jev service. Relative to Jev-Mem, the intervention moves from constructing and traversing stored relations to building query-time evidence chains through lexical expansion.

\FloatBarrier
\subsection{Decision Models in Agent Learning}
\label{sec:downstream-learning}

\takeaway{obs:learner}{A decision model can guide another learner without itself being trained.}
A fixed decision service can also change an agent's exploration, rewards, or updates during learning. Any resulting benefit may persist in the trained agent after the service is removed. This differs from fitting a replacement decision model to service outputs, as in Appendix~\ref{sec:fitting}, where training changes the model that will answer future decision requests. The relevant distinction is which model learns from the service and what it learns to do.

\citet{jevrl} use Jev as an action-policy reference, an exploration source, an auxiliary reward assessor, and a replay-priority signal. These interventions change an agent's data or updates, so their success is measured in the learned agent. AutoTrust instead fits a replacement decision model to Jev's probability outputs, making reproduction of the teacher's predictions a training target.

The interventions have different semantics. Policy assistance supplies actions that can be imitated or compared with the learner's actions by a critic. Exploration assistance replaces some random actions with Jev suggestions. Reward assistance maps rubric judgments to bonuses or gates. Replay assistance reweights stored experience using Jev judgments. The reported replay variant omits importance correction and therefore intentionally changes the effective training weighting. None of these roles makes an ordinal judgment a validated cardinal Q-value merely by converting its label to a number.

Information access also matters: Jev can receive structured descriptions beyond the learner's own observation, and the trained learner can later be deployed without Jev or its auxiliary rewards. The paper's MiniGrid and Atari experiments provide evidence for particular learning interventions, with three seeds; replay assistance does not uniformly help. The resulting gains or losses concern how the service changes learning. They cannot be attributed to the quality of its decision probabilities without separating added information, changed exploration, and changed objectives.

\paragraph{Joint learning changes the decision model as well as the agent.}
Dyad spans both levels in a different way \citep{dyad}. Its language model retains token generation and switches to a typed action distribution at designated interaction positions. The model aligns state and action representations through projector training, then applies environment-reward policy optimization over token and action decisions. Its action encoder may be updated alone or jointly with the state encoder. Action features are reusable only while their environment context and descriptions remain unchanged. Thus the learned architecture belongs to model construction, while the interleaving of language and environment actions defines the downstream workflow; it is not simply a frozen service plugged into another learner.

\FloatBarrier
\section{From Local Probabilities to Workflow Reliability}
\label{sec:evaluation}

These workflow roles call for separate tests of candidate probabilities, selected actions, and complete task outcomes. Accurate probabilities can yield a poor action under an unsuitable decision rule; a correct local action can still fail to secure task success. Our experiment measures selections, leaving calibration and downstream utility unresolved. The studies below identify the evidence needed to connect these levels.

\subsection{Local Probability Quality and Decision Utility}

\takeaway{obs:probability}{Calibration, distribution fidelity, and decision utility are distinct.}
Ordinary calibration compares reported probabilities with outcome frequencies among cases receiving the same predicted probabilities. Subgroup calibration additionally requires this agreement within specified groups. Distribution fidelity instead asks whether the complete prediction matches the conditional target for each input. Neither calibration nor fidelity alone specifies the action taken or the cost of an error. A decision rule converts probabilities into behavior, so a utility improvement can arise from a better distribution, a better rule, or both. Comparisons must separate these changes before attributing a gain to learning.

\paragraph{Known distributions separate probability estimation from sampled correctness.}
ProbabilityBench supplies simulated problems with a known conditional distribution $p^*$ given the visible input \citep{probabilitybench}. This permits direct comparison with $p^*$, rather than treating one sampled answer as the complete probability target. Accuracy evaluated in expectation under these conditional targets is bounded by the Bayes accuracy for the same visible information. Accuracy against sampled test labels can exceed that reference through label-sampling noise. For a predicted distribution $q$, the input-averaged $\mathrm{KL}(p^*\Vert q)$ equals excess expected log loss relative to $p^*$. A constant class prior satisfies ordinary calibration while ignoring input-specific information, and need not be calibrated within subgroups. Finite-bin estimates can also hide canceling errors or reflect sampling noise. Exact uncertainty in a synthetic generator is therefore valuable evidence about probability estimation, but does not establish access to the true uncertainty of an arbitrary real task.

\paragraph{Additional computation can sharpen probabilities without improving calibration.}
SanSi provides a concrete example of why compute allocation and probability quality must be evaluated together \citep{sansi}. Its later loops continue to increase confidence even after most answers have stabilized. On the 9,645 answerable test items, the 1.4B model's expected calibration error rises from $0.082\pm0.010$ at loop three to $0.093\pm0.012$ at loop eight. These are means and standard deviations over three training seeds, using ten equal-width confidence bins. The comparison concerns different readout depths within each trained model. It shows that additional latent computation can worsen calibration even within the trained loop range, so a stopping rule needs evidence for the probability or decision objective it serves.

\paragraph{Utility also depends on the decision rule.}
TypedBench complements this view with policy-generated cases, controlled framing changes, and cost-sensitive evaluation \citep{typedbench}. Its calibration and stress-test protocols distinguish benign framing sensitivity from adversarial perturbation. A change from argmax to the Bayes action under a cost matrix changes the decision rule; the resulting utility improvement cannot all be credited to better calibration. Synthetic labels make some controls possible while limiting direct conclusions about unconstrained natural tasks.

\paragraph{Benchmark scores depend on their evaluation protocol.}
Published scores differ in both what they measure and how evaluation informs model development. JEVal supplies a benchmark as well as the SII model discussed in Appendix~\ref{sec:fitting}, but the SII release's separate evaluation suite does not reproduce that benchmark protocol \citep{jeval}. Inner Jev's Capability metric differs from ordinary accuracy, and its OpenDecisions results inform recipe selection \citep{inner}. LLM-as-Jev likewise uses public JevBench errors to guide data-mixture development \citep{llmasjev}. Metric definitions and evaluation feedback must therefore be accounted for before these scores can support a common ranking.

Data overlap and execution conditions further limit comparisons. FunDM's reproduction notes disclose overlap between a CLM post-training collection and its test collection. Local-model and hosted-API latency also reflect different hardware and network conditions. Generalization and efficiency claims therefore require controls beyond matching the reported metric.

\subsection{Composition and Input Sensitivity}

\takeaway{obs:composition}{Reliable isolated decisions need not remain reliable under composition.}
An isolated prediction tests one input and output interface. Composing predictions introduces dependencies; changing a request can alter evidence, eligible actions, or numerical computation. Interpreting sensitivity first requires specifying what should remain invariant. Our permutations preserve question content and candidate identities, whereas adding a candidate or revealing evidence can change the decision problem.

Coherence studies test probability relationships across related questions \citep{coherence}, while option-set studies test how predictions respond to changed alternatives \citep{iia}. Neither relationship follows from calibration on isolated questions.

\paragraph{Request context, candidate eligibility, and numerical precision.}
\citet{company} distinguishes changes to the questions bundled into a request, the evidence provided, and the eligible candidates. A model whose question compiler produces exactly the same input can be invariant for that reason, whereas a full-schema prompt exposes other questions to the decision computation. The Nimble intervention removes other questions from the prompt and eliminates the observed flips, but is not length-matched. Small Laya shifts largely disappear at higher precision or after rounding. The observed sensitivity can therefore reflect prompt construction or numerical precision as well as architectural dependencies. Conditional odds between retained alternatives separate genuine relative-score movement from the inevitable change in a softmax denominator. When a readout restricts the vocabulary, its allowed-label mass provides additional information lost by normalization.

\paragraph{From correct choices to correct sequences.}
Choice2Seq compares native candidate choice with independent-position and serial-prefix output interfaces on paired problems \citep{choice2seq}. Its serial correctness retention is $P(\text{serial correct}\mid\text{native correct})$, not agreement with the native answer or fidelity to a plan. Cases improved by the serial interface and the marginal accuracies are needed alongside this conditional statistic. Gold-copy and self-plan controls address different fidelity questions. Candidate restrictions can also change the available search space and number of calls, so a composition claim requires the relevant interface and information controls.

\paragraph{Robustness and program consequences.}
Robustness studies connect input sensitivity to the consequences of a valid but incorrect choice. DecisionRobust tests label-preserving perturbations and targeted prompt injection on initially correct decisions across text and vision \citep{decisionrobust}. The distinction between observation text and task instructions matters because changing task facts would change the decision problem itself. JevProbe traces the structured response through application logic and tests whether reachable lower-trust context can redirect a program decision while leaving task facts and code fixed \citep{jevprobe}. Together, these studies show why schema validity is insufficient for reliable execution. They identify failures of input handling and decision use, with modeled program consequences rather than measured real-world losses.

\paragraph{The released response can reveal more than the intended decision.}
TUPLE-Bench studies whether a response about an authorized coarse category reveals a protected finer mechanism in the model's private evidence \citep{tuplebench}. Choice-only, confidence, full-probability, and native-response releases are derived from the same saved response, separating release policy from model reruns. Its informed decoder receives the gold coarse category and two known fine-grained hypotheses; this is a passive closed-set audit, not an estimate of unrestricted real-world leakage. A separate legitimate consumer sees neither private evidence nor those hidden labels. The proposed workflow-conditioned probability quantization selects codewords that preserve a frozen consumer's actions, including clarification, with a native-response fallback when no match exists. It can preserve wrong actions as well as correct ones, so action preservation establishes neither improved task quality nor a confidentiality guarantee. The contribution concerns evaluation and release postprocessing, not a new decision-model training objective.

\subsection{Complete-Workflow Outcomes}

\takeaway{obs:workflow-evaluation}{Task-level gains depend on the surrounding policy and its total resource use.}

\citet{agents} separate changes to the action interface from changes to the decision model in ALFWorld. They compare the original baseline, an LLM using a Choice interface, and Jev using that interface. The first contrast examines the candidate-selection interface; the second compares decision components within it. Their AndroidWorld experiments compare baseline and Jev integrations without this three-way control. Across these workflows, local decision savings do not uniformly translate into better task outcomes. Action granularity also differs in one integration: FastAgent-J takes one tool action per step while its baseline can take multiple actions. The study thus provides evidence about particular integrations while illustrating the controls needed to interpret them.

JevAgentHarness distinguishes isolated multiple-choice accuracy, decisions on recorded agent states without execution, and actual closed-loop intervention \citep{harness}. Its closed-loop tool-selection comparison lets the LLM supply tool arguments and uses unthresholded replacement. That protocol differs from the confidence cascade proposed for deployment, so its outcomes cannot be attributed directly to that cascade. Scripted acceptable actions in DecisionGym and disjoint rollout seeds also do not, by themselves, establish generalization to unseen task distributions.

System1-Arena broadens the comparison across static decisions and closed-loop environments under task-specific shared harnesses \citep{systemonearena}. Candidate construction converts otherwise open-ended search, patches, or arguments into finite alternatives; the reported results are conditional on this interface. The static comparison uses 85,305 questions across 94 tasks, a subset of the released collection, and excludes its game and embodied questions. Reported static scores weight tasks equally, with source averaging within a task. Closed-loop scores weight environments equally. For real-time simulations, the environment advances using each model's measured median latency while retaining the prior action, rather than measuring each call's realized delay inside the simulation. Probability extraction also differs across native decision models, verbalized-probability APIs, and local label-logit models. These protocol choices enable useful controlled comparisons while limiting a universal ranking across interfaces, hardware, and workloads.

\paragraph{Adoption is a separate object of measurement.}
Ecosystem studies explain where decision models are used, but do not replace these capability and workflow evaluations. \citet{wild} analyze public Jev repositories by application domain and decision purpose. Purposes can overlap within a project, and repository counts or stars measure ecosystem activity rather than model quality or methodological novelty. \citet{typedaudit} audit early typed-decision-model evidence and connect it to label-probability readouts, reranking, calibration, and model cascades.

\section{Open Questions}
\label{sec:rethink}

The unresolved question is which design changes preserve correct choices across equivalent presentations. The comparisons below separate representation, learning, computation, and workflow effects so that any improvement can be attributed to a specified intervention.

\paragraph{Which improvements survive equivalent presentations and related decisions?}
Takeaways~\ref{obs:probability} and~\ref{obs:composition} distinguish local probability quality from consistency across requests. The main study finds substantially more Jev variation under reordering than under fixed-order repetition, without identifying the mechanism behind this difference. Independent studies examine disagreement between direct and hierarchical predictions \citep{coherence}, option-set dependence \citep{iia}, serial correctness retention \citep{choice2seq}, and the limited transfer of local savings to successful agent outcomes \citep{agents}. Together, they motivate testing whether a local probability improvement persists under transformations that preserve the intended decision problem and predicts utility in a fixed workflow. The intended invariance must be specified first. Independence of irrelevant alternatives is not universally appropriate, since an added option may provide meaningful context or change the task. Prompt serialization and precision require controls \citep{company}, while cacheability must be separated from set-dependent refinement \citep{jeva}.

\paragraph{When does the gradient estimator matter beyond the target and objective?}
Takeaway~\ref{obs:credit} separates statistical supervision from the variable receiving gradient credit. Replacing observed labels with teacher distributions changes the target; adding output noise can replace an unperturbed score with its expectation under perturbation. An estimator comparison therefore requires a fixed objective and perturbation distribution, with pathwise differentiation and score-function credit compared where both apply. Matching the backbone, prompt, candidates, examples, adaptation budget, and validation split would isolate this question, with frozen and temperature-calibrated readouts as starting controls. Whether a gain survives common calibration and probability postprocessing would help establish its source. A negative result would delimit the value of a specified stochastic update without resolving how proprietary Jev is trained.

\paragraph{Which aspects of reasoning transfer into a direct readout?}
Takeaways~\ref{obs:targets} and~\ref{obs:computation} distinguish the information supplied during learning from the computation available at inference. JEVal reports that reasoning-to-readout transfer differs between evidence-weighing and computation-heavy tasks \citep{jeval}; earlier System~2 distillation also documents task-dependent transfer \citep{system2}. Conversely, theoretical work shows that some computations can be internalized under particular architectures and curricula \citep{internalize}. The unresolved empirical question concerns the interaction of training support, novel rule composition, and dependency depth. Correct-label supervision, additional data, intermediate-state hints, and explicit-reasoning controls would help distinguish imperfect targets and learning limitations from residual computational demands. Failure on a task family alone does not prove an impossibility of single-pass reasoning.

\paragraph{When does a better decision model improve the complete workflow?}
Takeaways~\ref{obs:intervention}--\ref{obs:learner} identify where a model intervenes; Takeaway~\ref{obs:workflow-evaluation} shows why that intervention matters for evaluation. A comparison within a fixed workflow can replace only the decision component while holding candidate construction, retrieval, budgets, fallback rules, and answer generation constant. A complementary comparison retains the same model while changing its role or control policy. Jev-Mem motivates separating write control from read control; run selection requires matching candidate-run generation; LAVOIR requires accounting for newly acquired information and the cost of questions. Crossing model and workflow choices can reveal whether local probability improvements transfer to task outcomes and how that transfer depends on the surrounding policy. Existing evidence does not establish that either model quality or workflow design universally dominates.

These factors can interact. Establishing a structural advantage requires identifying both the intervention and the conditions under which its benefit survives.

\begin{samepage}
\section{Survey Synthesis}
\label{sec:conclusion}

A candidate-distribution interface accommodates different representations, objectives, gradient estimators, inference procedures, and downstream roles. The main study shows that this shared interface does not ensure preservation of correct choices under reordering. Explaining the observed differences requires controlled comparisons under common information and resource budgets, followed by tests of whether local improvements benefit the workflow the model controls.

\end{samepage}

\section{Extended Public Implementation Inventory}
\label{sec:inventory}
Table~\ref{tab:inventory} summarizes additional implementations and the mechanisms documented in their primary sources as of October~19, 2026. Public artifacts do not by themselves establish independent reproduction.

\begingroup
\small
\setlength{\tabcolsep}{3pt}
\renewcommand{\arraystretch}{1.18}
\begin{longtable}{@{}L{2.85cm}L{5.3cm}L{5.35cm}@{}}
\caption{Additional implementations and documented mechanisms.}\label{tab:inventory}\\
\toprule
\textbf{Project / primary source} & \textbf{Documented mechanism} & \textbf{Scope and distinctions}\\
\midrule
\endfirsthead
\multicolumn{3}{l}{\small Table~\thetable\ continued}\\
\toprule
\textbf{Project / primary source} & \textbf{Documented mechanism} & \textbf{Scope and distinctions}\\
\midrule
\endhead
\bottomrule
\endfoot
\multicolumn{3}{l}{\textit{Frozen readout and execution recipes}}\\*
\href{https://github.com/TheoLeeCJ/SemIf-OpenJev}{SemIf / former OpenJev} & Restricted answer-label probabilities; workload-specific temperature fitting & Runtime code; explicitly an interface pattern, not TypeSafe's training recipe\\
\href{https://github.com/ikermoel/open-alternative-jev}{open-alternative-jev} & Multiple packed readout positions; exact-logit and top-$k$ logprob backends & Runtime variants differ; some use constrained one-token generation\\
\href{https://github.com/blockbrain-ai/cygnet-recipe}{Cygnet} & Frozen Gemma-4-12B-it; label probabilities, fitted temperature and large-menu grouping & One output token is emitted; no weight adaptation\\
\href{https://github.com/krishna-gogineni-765/deck31b}{deck31b} & Frozen Gemma-4-31B-it for candidate selection & Uses pretrained weights without decision-specific adaptation\\
\href{https://github.com/kshetrajna12/reflex}{reflex} & Frozen Qwen3.5-4B with averaging over two option orders & Tested adapters degraded out-of-distribution judgments; the default model remains frozen\\
\href{https://github.com/nokia-applied-research/AnyJev}{AnyJev Decider L0} & Cyclic label rotations and label-prior correction & Full rotation costs $K$ prefills for $K$ candidates; no weight adaptation\\
\multicolumn{3}{l}{\textit{Adaptation and teacher-based variants}}\\*
\href{https://github.com/strands-labs/strands-decider}{Strands Decider} & v1: Qwen3.5-2B-Base, LoRA and pointer head; supervised targets, model averaging and calibration & Training loop and inference sources available; optional KL branches do not prove their use in every released variant\\
\href{https://github.com/bespokelabsai/nimble}{Nimble} & Qwen3.5-9B LoRA; hard-label CE over candidate logits; paired examples stay in one split & ``Contrastive'' describes paired data, not an InfoNCE objective; stored Jev probabilities are not the targets of this loss\\
\href{https://github.com/InternLM/Intern-Decision}{Intern-Decision} \citep{interndecision} & Native logits before decision placeholders; full-vocabulary CE at those shifted positions; frozen vision encoder and projector & Technical report and training entry points available; original training data and private calibration splits are not released\\
\href{https://github.com/OmniJev/OneJev}{OneJev} & Full language adaptation with frozen vision; candidate-slot soft-target CE plus Brier loss & Public data are a subset; not the OmniJev model evaluated by FastJEV\\
\href{https://github.com/jaredpalmer/kev}{Kev} & CE-trained pointer head; smaller models use LoRA, 27B uses full adaptation & Training/inference/evaluation sources and model links; hybrid models use separate causal question rows with serving state reuse\\
\href{https://github.com/allebee/jevk5}{JevK5} & Qwen3.5-4B and Qwen3.5-9B LoRA, teacher-derived data, letter readout and temperature & Up to 16 choices per pass; tournaments for larger menus\\
\href{https://github.com/crh225/plumb}{Plumb} & Teacher-generated hard labels plus public-label replay; LoRA CE & Pipeline described; related to JevK5; hard labels do not imply soft-target KL\\
\href{https://huggingface.co/HopitAI/hopper}{Hopper} & LoRA and mixed synthetic, teacher-reviewed and public supervision & Model card describes multiple backbone variants\\
\href{https://huggingface.co/morriszjm/Tacit-9B}{Tacit / AnyJev} & Same-base task/reasoning-label self-distillation; optional thinking router & Distinct from training-free Decider L0\\
\href{https://huggingface.co/autotrust/JEV-27B}{AutoTrust-JEV-27B} & Jev~1.13 distributions mixed with programmatic and downweighted placeholder targets; KL + ordinal RPS & Detailed recipe, weights and inference; teacher agreement is not correctness\\
\href{https://huggingface.co/caiovicentino1/Eikos-27B}{Eikos} & External-teacher soft CE, LoRA and auxiliary rationale loss & Model-card evidence; no inference-time reasoning requirement inferred from that loss\\
\href{https://huggingface.co/torchcast-ai/torchcast-decision-12b}{Torchcast} & Gold labels and larger-model probability targets; LoRA & Model card and serving release; public benchmark used in model/configuration selection\\
\href{https://huggingface.co/EldanRing/Winnow-12B}{Winnow} & Trained LoRA with synthetic, teacher and labeled data & Private training pipeline; exact loss undisclosed; inference code available\\
\href{https://huggingface.co/chinhnc/Quyet-1.0-Large}{Quyet-1.0-Large} & Merged LoRA; letter readout and per-type temperature & Description applies to the Large variant\\
\href{https://huggingface.co/flymy-ai/decision-fast-preview}{decision-fast-preview} & Trained adapter, structural tokens and pointer head & Adapter, head and inference code released; mixed corpus not redistributed\\
\href{https://arxiv.org/abs/2609.36115}{Koa-decision} & Reserved atomic label tokens with randomized assignment; single-position CE & Training recipe described in the paper\\
\multicolumn{3}{l}{\textit{Policy recipes and version-specific training}}\\*
\href{https://github.com/NandhaKishorM/laya}{Laya} & Bidirectional decision model; soft CE and perturbed-logit policy updates & Chinese-Jev explicitly inherits this loss; distinct from reasoning-token RL\\
\href{https://github.com/Mapika/decider}{decider} & 2B v10: PPO environment-action policy, terminal returns and auxiliary log-score belief loss & Documented 4B/35B omit this RL stage; stock decider-chat is frozen; RL loop not in release package\\
\href{https://github.com/PostHog/jeeves}{Jeeves} & Pointer head; SFT, reasoning CISPO and diffusion drafting & Public trainer; correct-answer-probability reward with length penalty is not a proper score\\
\href{https://github.com/Shanghua-Gao/RSI-Jev}{RSI-Jev} & Versioned SFT, averaging and early-exit variants; retained ranking RL uses NDCG@5 & Task-specific ranking reward; recipes and benchmark-selection exposure vary by version\\
\multicolumn{3}{l}{\textit{Task-oriented model extensions}}\\*
\href{https://github.com/moganai/lavoir}{LAVOIR} \citep{lavoir} & Extends Laya with decision CE and value-of-information regression; interaction in Appendix~\ref{sec:clarification} & Final broad release sets inherited RL weight to zero; also contributes a downstream policy\\*
\href{https://github.com/TianyuCodings/NanoJev}{NanoJev} & Complete-question CE on four games & Supervised game-decision model; distinct from uses of NanoJev as a simulator\\
\end{longtable}
\endgroup

\begin{samepage}
\paragraph{Documented readout, undisclosed training objective.}
\href{https://huggingface.co/Cloudflare/clef}{Cloudflare Clef} attaches a small transformer joint schema head to Qwen3.8-27B. Its card describes all-question probabilities from one forward pass, an expected-value scalar score, and backbone, head and inference artifacts, but not the exact post-training objective. \href{https://huggingface.co/perplexity-ai/pplx-decider-v1.1-27b}{pplx-decider-v1.1-27b} uses a separate $255\times5120$ readout, noncausal full-attention layers with unchanged linear-attention behavior, and saved temperature calibration. Its training loss and update method remain unspecified. These interfaces alone establish neither RL nor supervised training.

\end{samepage}

\paragraph{Additional documented implementations.}
\href{https://github.com/mohit67890/imajev}{imajev} documents language LoRA with frozen vision and weight-space averaging of a hard-question adapter with its soft-target continuation.
\href{https://github.com/Octalab-Inc/jqv}{jqv} offers frozen-logit backends and trainable slot or pointer heads with LoRA and CE plus Brier loss.
\href{https://github.com/Heman10x-NGU/openJev-verdict-2.0}{openJev-verdict} describes a marker-pointer distribution head, an out-of-fold confidence head, and symmetric KL between option-shuffled training pairs.

Other designs vary the candidate representation.
\href{https://github.com/vinnylarouge/jevlike}{jevlike} learns option-conditioned context scoring with either a learned byte encoder or a frozen pretrained encoder.
\href{https://github.com/malevrigns/agent-jev}{agent-jev} combines independent candidate paths with a permutation-equivariant scoring head and documents soft CE plus Brier training for AgentJev-0.6B.
\href{https://huggingface.co/fastino/GLiNER2.5-Decide}{GLiNER2.5-Decide} publishes supervised fine-tuning formats and procedures, but the objective used for its released weights remains unspecified.
\href{https://github.com/us/jev-local}{jev-local} offers an optional frozen-model log-probability scorer and a deterministic default stub; head training is explicitly deferred.

\end{document}